\documentclass[11pt]{article}

\usepackage[final]{acl}

\usepackage{times}
\usepackage{latexsym}

\usepackage[T1]{fontenc}

\usepackage[utf8]{inputenc}

\usepackage{microtype}

\usepackage{inconsolata}

\usepackage{graphicx}

\usepackage{algorithm}
\usepackage{algorithmic}
\usepackage{amsmath}
\usepackage{amsfonts}
\usepackage{booktabs}
\usepackage{colortbl}
\usepackage[table]{xcolor}
\usepackage{tabularx}
\usepackage{multirow}
\usepackage{makecell}
\usepackage{diagbox}
\usepackage{enumitem}
\usepackage{cuted}
\usepackage{hyperref}
\usepackage{tcolorbox}

\newcommand{\adjustimg}{
  \hspace*{\dimexpr\evensidemargin-\oddsidemargin}%
}
\newcommand{\centerimg}[2][width=\textwidth]{
  \makebox[\textwidth]{\adjustimg\includegraphics[#1]{#2}}%
}

\title{Mitigating Cross-Image Information Leakage in Multi-Image Understanding with Large Vision-Language Models}

\author{First Author \\
  Affiliation / Address line 1 \\
  Affiliation / Address line 2 \\
  Affiliation / Address line 3 \\
  \texttt{email@domain} \\\And
  Second Author \\
  Affiliation / Address line 1 \\
  Affiliation / Address line 2 \\
  Affiliation / Address line 3 \\
  \texttt{email@domain} \\}

\author{
Yeji Park\textsuperscript{1} \quad
Minyoung Lee\textsuperscript{1} \quad
Sanghyuk Chun\textsuperscript{2}\thanks{Works done at NAVER AI Lab.} \quad
Junsuk Choe\textsuperscript{1}\thanks{Corresponding author.}
\\
\textsuperscript{1}Sogang University \quad
\textsuperscript{2}Princeton University
}

\begin{document}
\maketitle
\begin{strip}
  \vspace{-58pt}

    \noindent\centerimg[width=\linewidth]{Figures/Introduction.pdf}
 
  \fontsize{10pt}{12pt}\selectfont
    Figure 1: Illustration of \textit{cross-image information leakage} in LVLMs in single-image vs. multi-image settings. Under the multi-image setting, the model entangles visual features across images. As a result, the model incorrectly selects the option that combines objects from both images, rather than the object present only in the target image.
  \label{fig:introduction}

\end{strip}

\setcounter{figure}{1}

\begin{abstract}
Large Vision-Language Models (LVLMs) exhibit strong performance on single-image tasks.
However, their performance degrades significantly when handling multi-image inputs. 
While this degradation has been observed in prior work, its nature remains poorly understood.
We empirically observe visual elements from different images become entangled in the model’s representations and responses.
We refer to this phenomenon as \textit{cross-image information leakage}.
To address this issue, we propose \textbf{\textsc{FOCUS}}, a training-free and architecture-agnostic method.
\textsc{FOCUS} masks all but one image with random noise, guiding the model to focus on the single clean image.
This process is applied across the target images to obtain logits under partially masked contexts.
These logits are aggregated and then refined using a noise-only reference input, which suppresses the leakage and yields more accurate outputs.
\textsc{FOCUS} consistently improves performance on diverse multi-image benchmarks. We further show that \textsc{FOCUS} generalizes to video understanding, extending its applicability beyond static multi-image inputs.
This demonstrates that \textsc{FOCUS} offers a general solution for enhancing multi-image reasoning without additional training or architectural modifications. Source code is available at \url{https://github.com/yejipark-m/FOCUS}.
\end{abstract}  
\section{Introduction}

Large Vision-Language Models (LVLMs) are designed to jointly understand visual and textual information~\cite{Chen_2024_CVPR,achiam2023gpt, bai2025qwen2}. They excel at vision-language (VL) tasks~\cite{li2024allava,daxberger2025mm}, including Visual Question Answering (VQA)~\cite{antol2015vqa} and Image Captioning~\cite{herdade2019image}. 

Although LVLMs have shown strong performance on VL tasks with a single image, real-world use cases often go beyond this single-image assumption. Users naturally provide multiple images at once and expect models to integrate information across all of them, such as comparing scenes and resolving cross-image references to produce a coherent response~\cite{jiang2024mantis}. To this end, several studies train models with interleaved image-text sequences to explicitly enhance reasoning across multiple images~\cite{awadalla2023openflamingo, lin2024vila, sun2024generative, jiang2024mantis}. 

However, recent studies have revealed that LVLMs still struggle in such settings, showing a notable performance drop when multiple images are provided~\cite{wang2025muirbench, zhang-etal-2025-weaving}. This suggests that the underlying cause of this degradation cannot be fully attributed to insufficient training data. Yet, why this performance degradation occurs remains largely unexplored.

\begin{figure*}[t]
  \includegraphics[width=\textwidth]{Figures/Empirical.pdf}
  \caption {Illustration of cross-image information leakage across different image pairs. Given the same target image (red border), the baseline LVLM consistently selects the incorrect merged caption that combines content from both images (Failure Cases), regardless of which non-target image is present. FOCUS mitigates this confusion and recovers the correct answer in both cases (Corrected Results).}
  \label{fig:empirical}
\end{figure*}
We observe that LVLMs tend to exhibit a phenomenon we term \textit{cross-image information leakage}, where models mix visual cues across inputs rather than interpreting each image independently.
As illustrated in Figure~\hyperlink{fig:introduction}{1}, a model answers correctly with a single image, but fails when two images are provided simultaneously. Specifically, in the multi-image setting, the probability of ``\textit{yellow tulips}'' (the correct option) decreases. Instead, the model mistakenly selects ``\textit{yellow tulips and red roses}''. This answer merges visual content from two separate images into a single, incorrect response. This suggests that current LVLMs cannot reliably isolate and reason over visual information from multiple images independently. 

Moreover, this failure is not specific to a particular image pair. In Figure~\ref{fig:empirical}, when the non-target image changes from blue irises to yellow sunflowers, the model predicts its incorrect answer from ``\textit{yellow tulips and blue irises}'' to ``\textit{yellow tulips and sunflowers}'', consistently incorporating content from whichever non-target image is present.

Our analysis shows that cross-image information leakage is closely related to multi-image understanding degradation. (\S\ref{sec:leakage_motiv}).
Under multi-image inputs, models show an increased tendency to select captions that merge content from both the target and non-target images, leading to substantial performance degradation.
To mitigate this, one naïve solution is to process each image independently, given that models perform well on single images. Yet this fails to capture inter-image relationships that are essential for multi-image reasoning (\S\ref{sec:motiv_limitations}).

To this end, we propose \textsc{FOCUS}, a training-free method that mitigates cross-image information leakage while preserving inter-image relationships.
Our key idea is to leverage the observation that LVLMs reason well over a single image. We process all images together to preserve inter-image context, while guiding the model to focus on one image at a time.
Specifically, we mask all but one image with random noise, prompting the model to concentrate on the single clean image (\S\ref{sec:method}).
\textsc{FOCUS} applies this masking to each image in turn to extract the focused logits, and aggregates the resulting output logits while preserving the positional context of the images.

We validate \textsc{FOCUS} using diverse LVLM families across five multi-image benchmarks: Winoground~\cite{thrush2022winoground}, VisMin~\cite{awal2024vismin}, Mantis-Eval~\cite{jiang2024mantis}, MuirBench~\cite{wang2025muirbench} and MIRB~\cite{zhao2024benchmarking}. \textsc{FOCUS} achieves consistent improvements, with the best gains of up to  +18.8 Image and +16.8 Group score on Winoground, +44.37 Image and +41.9 Group score on VisMin, +5.5\%pts accuracy on Mantis-Eval, +3.2\%pts on MuirBench, and +1.7\%pts on MIRB.
Beyond static multi-image settings, \textsc{FOCUS} further generalizes to video understanding~\cite{zhang2024vinoground}. All gains are achieved without any additional training or architectural changes, highlighting the effectiveness and generalization ability of our method.

\begin{figure*}[t]
  \includegraphics[width=\textwidth]{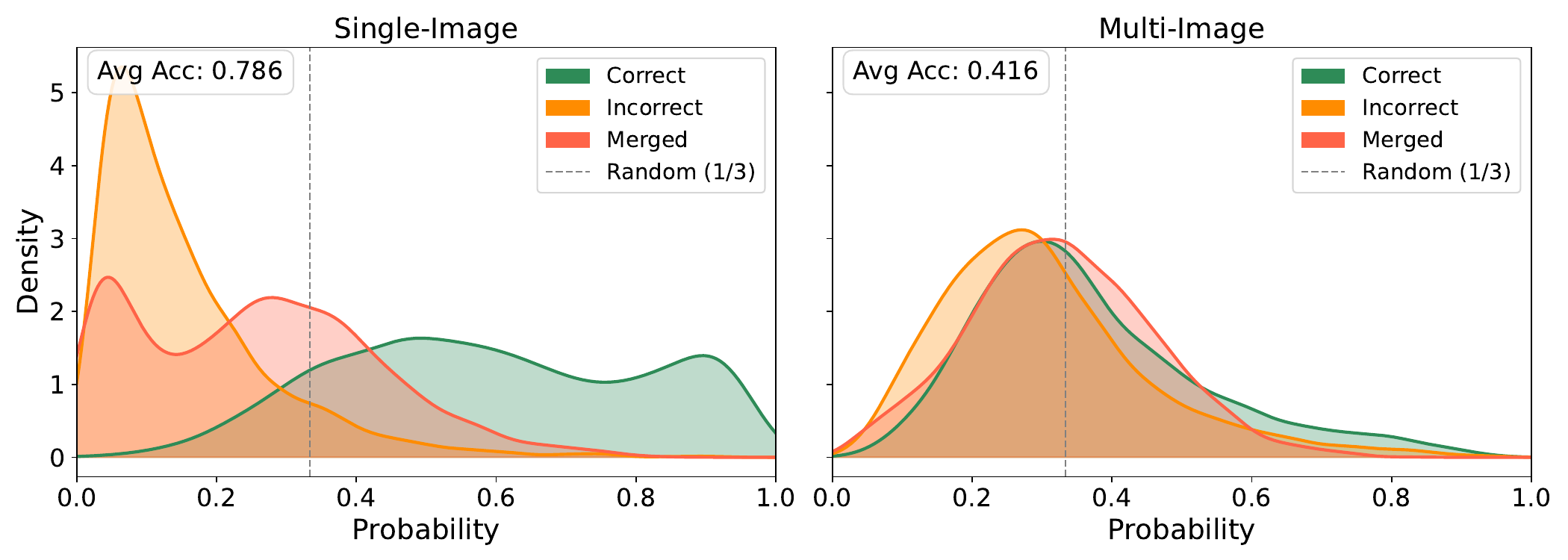}
  \caption {Estimation of probability density function of Qwen2.5-VL 3B predictions over three answer options. We evaluated on 500 image pairs from VisMin~\cite{awal2024vismin}. In the single-image setting, the correct option is favored. In the multi-image setting, all distributions located their center near random chance (the dotted line).}
  \label{fig:motivation_analysis}
\end{figure*}
\section{Motivation}

\subsection{Cross-image Information Leakage}
\label{sec:leakage_motiv} 
We observe that the model exhibits a notable limitation when processing multiple images simultaneously, namely \textit{cross-image information leakage}.

This issue is closely related to how visual inputs are embedded and processed within an LVLM. Most LVLMs handle multiple images as a sequence of visual tokens, which are concatenated with text tokens and processed through the attention mechanism. 
During this process, the tokens interact strongly with each other due to \textit{inter-image causal attention}~\cite{tian2025identifying}. This can cause the attention mechanism to mix latent representations across images and the model entangles visual semantics that should remain distinct.

We empirically analyze this using 500 image pairs from \citet{awal2024vismin}. For each pair, we provide three options: a correct caption for the target image, an incorrect caption from the other image, and a merged caption combining both (see Figure~\hyperlink{fig:introduction}{1}, options (a)-(c)). We estimate the probability density function of model predictions over these three options using kernel density estimation. 

As in Figure~\ref{fig:motivation_analysis}, in the single-image setting, the correct caption distribution is skewed toward higher probabilities, while the incorrect and merged caption distributions concentrate at lower probabilities. 
However, in the multi-image setting, the probability of selecting the merged caption increases. For example, in Qwen2.5-VL-3B model, all three distributions center near random chance and become nearly indistinguishable from one another.

These results suggest that the model loses its ability to selectively attend to the target image, making accurate multi-image reasoning substantially less reliable.
We further verify that this phenomenon persists across larger model scales, including Qwen2.5-VL-7B and Qwen3-VL-8B, though the degree of degradation vary with model capacity (provided in Appendix~\S\ref{app:leakage_analysis_scaleup}).

\subsection{Limitation of Independent Inferences}
\label{sec:motiv_limitations}
A straightforward approach to mitigate cross-image information leakage is to perform inference on each image independently and then aggregate the results. However, even an optimal aggregation scheme cannot capture \textit{inter-image relationships}, which are crucial for multi-image reasoning. Instructions such as ``\textit{Compare the second image with the first one}'' or ``\textit{Describe what is common across all images}'' inherently require reasoning about relationships between images, and such cross-referential queries are frequent in multi-image tasks.

One natural extension of this approach is to encode inter-image relationships explicitly through textual prompts. For example, one can generate an independent caption for each image and then concatenate these descriptions alongside positional information (\emph{e.g.}, ``The first image shows~\ldots, the second image shows~\ldots'') as 
context for a final inference step.  However, we experiment with this approach and find it to be ineffective. This limitation is further analyzed in \S\ref{sec:Comparison}.

\begin{tcolorbox}[title={\fontsize{11}{13}\selectfont Observations}, boxrule=0.5pt, arc=3pt, left=6pt, right=6pt, top=4pt, bottom=4pt, fontupper={\fontsize{11}{13}\selectfont}]
    \begin{itemize}[leftmargin=*, nosep]
    \item Cross-image information leakage is closely related to multi-image understanding degradation, where models increasingly confuse visual content across images and produce incorrect responses.
    \item Processing each image independently fails to resolve this issue, as it sacrifices inter-image relationships essential for multi-image reasoning.
    \end{itemize}
\end{tcolorbox}

Motivated by these observations, we aim to design a method that (i) jointly captures 
the structure and relationships among multiple images within a single, unified 
reasoning process, and (ii) mitigates cross-image information leakage. We address 
this challenge with \textsc{FOCUS}, described next.
\section{Method}
\label{sec:method}
\begin{figure}[t]
  \centering
  \includegraphics[width=\columnwidth]{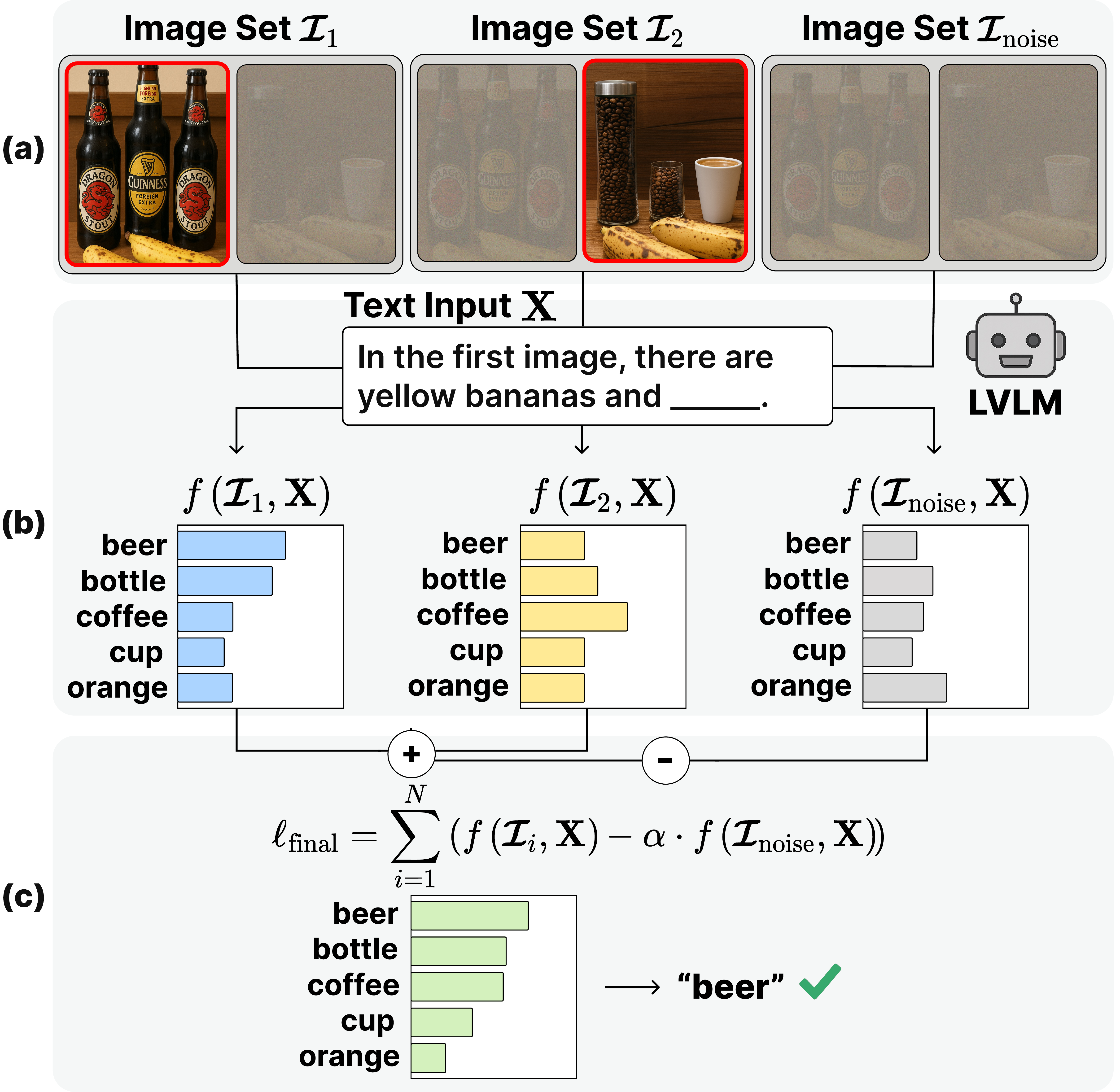}
  \caption{Overview of FOCUS. It consists of three main steps: (a) visual masking, (b) image-wise focused inference, and (c) contrastive aggregation.}
  \label{fig:2_method}
\end{figure}

Our method is a training-free approach that enables LVLMs to concentrate on one image at a time while preserving the positional structure of multi-image inputs. Our method consists of three steps: (a) visual masking, (b) image-wise focused inference, and (c) contrastive aggregation. A schematic overview of the proposed framework is shown in Figure~\ref{fig:2_method}.

\textbf{Visual Masking (Figure~\ref{fig:2_method}a).}
First, we prepare masked image inputs via noise injection with scale $\lambda$. Let $N$ be the number of input images and $\phi_v(\cdot)$ denote the visual encoder that maps an image to its token embeddings. For each target image $I_i$, we corrupt all other images $I_j$ ($j \neq i$) into $\tilde{I}_j$, leaving $I_i$ clean. This produces $N$ partially masked inputs and one fully masked reference:
\begin{equation}
\begin{aligned}
\mathcal{I}_i = [v'_1, \dots, v_i, \dots, v'_N], \\
\mathcal{I}_{\text{noise}} = [v'_1, v'_2, \dots, v'_N],
\end{aligned}
\end{equation}
where $v_i=\phi_v(I_i)$ and $v'_i=\phi_v(\tilde{I}_i)$ denote the clean and noise-corrupted visual embeddings, respectively. 
Each $\mathcal{I}_i$ allows the model to focus solely on $I_i$, and $\mathcal{I}_\text{noise}$ serves as a noise-only reference.

\textbf{Image-wise Focused Inference (Figure~\ref{fig:2_method}b).}
For each partially masked image input $I_i$, only the $i$-th image remains unmasked. The original image order is preserved to retain positional semantics. We compute the logit distribution $\ell_i$ for each $i$, where $f$ denotes the model, $\mathbf{X}$ denotes the text input:
\begin{equation}
\ell_i = f(\mathcal{I}_i, \textbf{X}).
\end{equation}

Figure~\ref{fig:2_method}(b) depicts the different logit distributions $\ell_i$ for $i = 1, \dots, N$ obtained via focused inference. For example, blue distribution (left) corresponds to the model focusing on the first image, while the yellow one (middle) represents focus on the second image (i.e., the clean image in $\mathcal{I}_2$).
We also compute a noise reference logit distribution using the fully noise-masked input:
\begin{equation}
\ell_{\text{noise}} = f(\mathcal{I}_{\text{noise}}, \textbf{X}).
\end{equation}
The gray logit distribution (right), which contains no information from either the first or the second image, serves as a reference for isolating irrelevant visual content during contrastive aggregation.

\textbf{Contrastive Aggregation (Figure~\ref{fig:2_method}c).}
Each $\ell_i$ contains useful signals from the clean image $I_i$, but also includes residual side-effects. For instance, in Figure~\ref{fig:2_method}(b), irrelevant token like \textit{orange} has nonzero probability by the noise masks. To suppress these residuals, we subtract the noise reference logit $\ell_{\text{noise}}$ from $\ell_i$ and aggregate the results:

\begin{equation}
\label{eq:f_final}
\ell_{\text{final}} = \sum_{i=1}^{N} \left(\ell_i - \alpha \cdot \ell_{\text{noise}}\right),
\end{equation}
where $\alpha$ is a hyperparameter tuned via validation to appropriately weight the subtraction. The final output token is then sampled from the aggregated logits $\ell_{\text{final}}$.
Figure~\ref{fig:2_method}(c) illustrates this step: the final green logit distribution, computed using Equation~\eqref{eq:f_final}, correctly identifies \textit{beer} as the answer.

The proposed method effectively suppresses cross-image information leakage while preserving positional structure, without requiring any model training or architecture modification. It enables LVLMs to perform multi-image reasoning tasks better in a training-free, generalizable manner.

\section{Experiments}
\begin{table}[t]
\centering
\small
\renewcommand{\arraystretch}{1.2}
\setlength{\tabcolsep}{3pt}
\begin{tabular}{l|ccc|ccc}
\toprule
\textbf{Model Size} & \multicolumn{3}{c|}{$<$\textbf{7B}} & \multicolumn{3}{c}{$\geq$\textbf{7B}} \\
\midrule
\diagbox[innerwidth=1.5cm]{\textbf{Family}}{\textbf{Score}} & \textbf{T} & \textbf{I} & \textbf{G} & \textbf{T} & \textbf{I} & \textbf{G} \\
\midrule
InternVL3
& 47.25 & 6.25 & 4.50
& \textbf{70.75} & 40.75 & 34.25 \\
\cellcolor{gray!10}+\textsc{FOCUS}
& \cellcolor{gray!10}47.25 & \cellcolor{gray!10}\textbf{27.25} & \cellcolor{gray!10}\textbf{19.75}
& \cellcolor{gray!10}69.25 & \cellcolor{gray!10}\textbf{42.25} & \cellcolor{gray!10}\textbf{35.75} \\
\midrule
Qwen2.5-VL
& 56.25 & \textbf{36.75} & 26.00
& 74.50 & 39.75 & 34.00 \\
\cellcolor{gray!10}+\textsc{FOCUS}
& \cellcolor{gray!10}56.25 & \cellcolor{gray!10}36.50 & \cellcolor{gray!10}26.00
& \cellcolor{gray!10}74.50 & \cellcolor{gray!10}\textbf{58.50} & \cellcolor{gray!10}\textbf{50.75} \\
\midrule
LLaVA-OV
& 3.25 & \textbf{20.00} & 0.25
& 76.75 & 36.75 & 33.00 \\
\cellcolor{gray!10}+\textsc{FOCUS}
& \cellcolor{gray!10}3.25 & \cellcolor{gray!10}19.25 & \cellcolor{gray!10}\textbf{0.75}
& \cellcolor{gray!10}76.75 & \cellcolor{gray!10}\textbf{48.50} & \cellcolor{gray!10}\textbf{42.25} \\
\bottomrule
\end{tabular}
\caption{Comparison on Winoground across model families and sizes with and without \textsc{FOCUS}. Metrics: T (Text), I (Image),
G (Group). Bold indicates improvement over the baseline.}
\label{tab:winoground}
\end{table}

\subsection{Benchmarks for Multi-image Tasks}
Multi-image understanding has been a fundamental challenge in visual reasoning \cite{suhr-etal-2017-corpus, suhr2018corpus}. We employ five evaluation benchmarks: Winoground~\cite{thrush2022winoground}, VisMin~\cite{awal2024vismin}, Mantis-Eval~\cite{jiang2024mantis}, MuirBench~\cite{wang2025muirbench} and MIRB~\cite{zhao2024benchmarking}.
We additionally evaluate \textsc{FOCUS} in video understanding and long-context settings using Vinoground~\cite{zhang2024vinoground} and MMLongBench~\cite{wang2026mmlongbench}, respectively.
More details and evaluation setup for each benchmark are provided in Appendix~\S\ref{app:benchmarks} and \S\ref{app:implementation_details}.

\subsection{Quantitative Results}

We evaluate the generalizability of the proposed method on three representative LVLM families: InternVL3 (2B, 8B)~\cite{Chen_2024_CVPR}, Qwen2.5-VL (3B, 7B)~\cite{bai2025qwen2}, and LLaVA-OneVision (0.5B, 7B)~\cite{li2024allava}. All hyperparameters are selected on the validation split. We report test split performance on each benchmark. To further validate generalizability across recent architectures, we additionally 
evaluate on Qwen3-VL (4B, 8B)~\cite{bai2025qwen3}.
Appendix~\S\ref{app:moe} extends this to larger models, including the MoE variant Qwen3-VL-30B-A3B and Qwen3-VL-32B.

\textbf{Results on Winoground.} Table~\ref{tab:winoground} presents the results on Winoground across three model families and scales, comparing baseline with the proposed \textsc{FOCUS} method. \textsc{FOCUS} consistently improves the Image Score (I) and Group Score (G). In InternVL3-2B, \textsc{FOCUS} increases the Image Score from 6.25 to 27.25. Notably, Qwen2.5-VL-7B achieves a substantial Image Score gain from 39.75 to 58.50. Overall, \textsc{FOCUS} tends to be effective across model families and scales, though the degree of improvement varies.

\begin{table}[t]
\centering
\small
\renewcommand{\arraystretch}{1.2}
\setlength{\tabcolsep}{3pt}
\begin{tabular}{l|ccc|ccc}
\toprule
\textbf{Model Size} & \multicolumn{3}{c|}{$<$\textbf{7B}} & \multicolumn{3}{c}{$\geq$\textbf{7B}} \\
\midrule
\diagbox[innerwidth=1.5cm]{\textbf{Family}}{\textbf{Score}} & \textbf{T} & \textbf{I} & \textbf{G} & \textbf{T} & \textbf{I} & \textbf{G} \\
\midrule
InternVL3        & 93.25 & 29.96 & 29.36 & 94.62 & \textbf{84.84} & \textbf{81.60} \\
\cellcolor{gray!10}+\textsc{FOCUS} & \cellcolor{gray!10}\textbf{93.34} & \cellcolor{gray!10}\textbf{74.33} & \cellcolor{gray!10}\textbf{71.26} & \cellcolor{gray!10}\textbf{94.70} & \cellcolor{gray!10}83.05 & \cellcolor{gray!10}80.83 \\
\midrule
Qwen2.5-VL        & \textbf{92.14} & 47.74 & 46.20 & 93.85 & \textbf{92.65} & 88.55 \\
\cellcolor{gray!10}+\textsc{FOCUS} & \cellcolor{gray!10}92.06 & \cellcolor{gray!10}\textbf{56.92} & \cellcolor{gray!10}\textbf{55.82} & \cellcolor{gray!10}93.85 & \cellcolor{gray!10}92.57 & \cellcolor{gray!10}\textbf{88.72} \\
\midrule
LLaVA-OV         & 20.98 & 9.26  & 1.55  & 93.43 & 62.99 & 60.68 \\
\cellcolor{gray!10}+\textsc{FOCUS} & \cellcolor{gray!10}20.98 & \cellcolor{gray!10}\textbf{27.01} & \cellcolor{gray!10}\textbf{8.42} & \cellcolor{gray!10}93.43 & \cellcolor{gray!10}\textbf{64.14} & \cellcolor{gray!10}\textbf{62.00} \\
\midrule
Qwen3-VL         & 94.20 & 90.68  & 87.35  & \textbf{94.88} & 93.24 & 90.34 \\
\cellcolor{gray!10}+\textsc{FOCUS} & \cellcolor{gray!10}\textbf{94.28} & \cellcolor{gray!10}\textbf{92.06} & \cellcolor{gray!10}\textbf{88.80} & \cellcolor{gray!10}94.79 & \cellcolor{gray!10}\textbf{93.50} & \cellcolor{gray!10}\textbf{90.51} \\
\bottomrule
\end{tabular}
\caption{Comparison on VisMin-Easy across model families and sizes with and without \textsc{FOCUS}. Metrics: T (Text), I (Image),
G (Group). Bold indicates improvement over the baseline.}
\label{tab:easy_set}
\end{table}

\begin{table}[t]
\centering
\small
\renewcommand{\arraystretch}{1.2}
\setlength{\tabcolsep}{3pt}
\begin{tabular}{l|ccc|ccc}
\toprule
\textbf{Model Size} & \multicolumn{3}{c|}{$<$\textbf{7B}} & \multicolumn{3}{c}{$\geq$\textbf{7B}} \\
\midrule
\diagbox[innerwidth=1.5cm]{\textbf{Family}}{\textbf{Score}} & \textbf{T} & \textbf{I} & \textbf{G} & \textbf{T} & \textbf{I} & \textbf{G} \\
\midrule
InternVL3        & \textbf{65.53} & 6.84  & 6.17  & \textbf{82.86} & 47.66 & 44.80 \\
\cellcolor{gray!10}+\textsc{FOCUS} & \cellcolor{gray!10}64.87 & \cellcolor{gray!10}\textbf{26.77} & \cellcolor{gray!10}\textbf{23.96} & \cellcolor{gray!10}82.54 & \cellcolor{gray!10}\textbf{52.66} & \cellcolor{gray!10}\textbf{48.89} \\
\midrule
Qwen2.5-VL        & 77.38 & 27.09 & 22.93 & 84.37 & 53.26 & 49.97 \\
\cellcolor{gray!10}+\textsc{FOCUS} & \cellcolor{gray!10}\textbf{78.20} & \cellcolor{gray!10}\textbf{28.86} & \cellcolor{gray!10}\textbf{25.29} & \cellcolor{gray!10}\textbf{84.47} & \cellcolor{gray!10}\textbf{61.44} & \cellcolor{gray!10}\textbf{56.99} \\
\midrule
LLaVA-OV         & 1.70  &\textbf{20.12} & 0.51  & \textbf{80.61} & 44.42 & 38.40 \\
\cellcolor{gray!10}+\textsc{FOCUS} & \cellcolor{gray!10}\textbf{2.20} & \cellcolor{gray!10}18.41 & \cellcolor{gray!10}\textbf{0.59} & \cellcolor{gray!10}80.44 & \cellcolor{gray!10}\textbf{57.89} & \cellcolor{gray!10}\textbf{49.40} \\
\midrule
Qwen3-VL         & 83.62 & 49.62 & 46.78  & \textbf{82.86} & 56.65 & 52.95 \\
\cellcolor{gray!10}+\textsc{FOCUS} & \cellcolor{gray!10}\textbf{84.11} & \cellcolor{gray!10}\textbf{58.31} & \cellcolor{gray!10}\textbf{54.32} & \cellcolor{gray!10}82.77 & \cellcolor{gray!10}\textbf{59.13} & \cellcolor{gray!10}\textbf{55.01} \\
\bottomrule
\end{tabular}
\caption{Comparison on VisMin-Hard across model families and sizes with and without \textsc{FOCUS}. Metrics: T (Text), I (Image),
G (Group). Bold indicates improvement over the baseline.}
\label{tab:hard_set}
\end{table}

\textbf{Results on VisMin-Easy.}
In Table~\ref{tab:easy_set}, \textsc{FOCUS} shows substantial improvements across model families. The gains are particularly pronounced in smaller models. InternVL3-2B improves its Image Score from 29.96 to 74.33 and Group Score from 29.36 to 71.26. For larger models, improvements are more modest due to stronger baselines.

\textbf{Results on VisMin-Hard.}
In Table~\ref{tab:hard_set}, baseline performance is comparatively lower, as these categories require more subtle cross-image discrimination. \textsc{FOCUS} still yields consistent improvements across model families. Qwen2.5-VL-7B improves its Image Score from 53.26 to 61.44 and Group Score from 49.97 to 56.99, while Text Score remains stable.
Notably, Qwen3-VL-4B achieves an improvement of +8.69 Image Score, confirming that \textsc{FOCUS} remains effective even on the latest models.

\begin{tcolorbox}[boxrule=0.5pt, arc=3pt, left=6pt, right=6pt, top=4pt, bottom=4pt]
{\fontsize{11}{13}\selectfont \textbf{Finding.} \textsc{FOCUS} improves Image Score (I) and Group Score (G), the metrics that directly reflect multi-image reasoning, while leaving Text Score (T) largely unchanged.}
\end{tcolorbox}

\begin{table}[t]
\centering
\small
\renewcommand{\arraystretch}{1.2}
\setlength{\tabcolsep}{2pt}
\begin{tabular}{l|ccc|ccc}
\toprule
\textbf{Model Size} & \multicolumn{3}{c|}{$<$\textbf{7B}} & \multicolumn{3}{c}{$\geq$\textbf{7B}} \\
\midrule
\diagbox[innerwidth=1.7cm]{\textbf{Family}}{\textbf{Dataset}} & \textbf{Mant.} & \textbf{Muir.} & \textbf{MIRB} & \textbf{Mant.} & \textbf{Muir.} & \textbf{MIRB} \\
\midrule
InternVL3
& 49.77 & 28.04 & 31.32
& 64.98 & 31.62 & 41.86 \\
\cellcolor{gray!10}+\textsc{FOCUS}
& \cellcolor{gray!10}\textbf{52.53} & \cellcolor{gray!10}\textbf{28.58} & \cellcolor{gray!10}\textbf{32.36}
& \cellcolor{gray!10}\textbf{65.44} & \cellcolor{gray!10}\textbf{31.88} & \cellcolor{gray!10}\textbf{42.48} \\
\midrule
Qwen2.5-VL
& 55.30 & 30.38 & 49.27
& 70.05 & \textbf{29.92} & 50.53 \\
\cellcolor{gray!10}+\textsc{FOCUS}
& \cellcolor{gray!10}\textbf{58.99} & \cellcolor{gray!10}\textbf{31.31} & \cellcolor{gray!10}\textbf{50.00}
& \cellcolor{gray!10}70.05 & \cellcolor{gray!10}29.73 & \cellcolor{gray!10}\textbf{52.30} \\
\midrule
LLaVA-OV
& 36.41 & 21.42 & \textbf{22.55}
& 57.14 & 28.73 & 48.02 \\
\cellcolor{gray!10}+\textsc{FOCUS}
& \cellcolor{gray!10}\textbf{41.94} & \cellcolor{gray!10}\textbf{24.62} & \cellcolor{gray!10}22.13
& \cellcolor{gray!10}\textbf{59.91} & \cellcolor{gray!10}\textbf{29.81} & \cellcolor{gray!10}\textbf{48.85} \\
\midrule
\textbf{Avg}
& 47.16 & 26.61 & 34.38
& 64.06 & 30.09 & 46.80 \\
\cellcolor{gray!10}+\textsc{FOCUS}
& \cellcolor{gray!10}\textbf{51.15} & \cellcolor{gray!10}\textbf{28.17} & \cellcolor{gray!10}\textbf{34.83}
& \cellcolor{gray!10}\textbf{65.13} & \cellcolor{gray!10}\textbf{30.47} & \cellcolor{gray!10}\textbf{47.88}\\
\bottomrule
\end{tabular}
\caption{Mantis-Eval, MuirBench and MIRB accuracy across
model families and sizes, with and without \textsc{FOCUS}. Bold indicates improvement over the baseline.}
\label{tab:mantis_muir_mirb}
\end{table}

\textbf{Results on Other Benchmarks.}
Table~\ref{tab:mantis_muir_mirb} summarizes the performance of six model variants across three benchmarks.

On \text{Mantis-Eval}, applying \textsc{FOCUS} leads to consistent accuracy gains across all settings. For instance, InternVL3-2B improves by +2.76\%pts, while the larger 8B model exhibits a smaller but positive effect. Similarly, Qwen2.5-VL-3B gains +3.69\%pts, and LLaVA-OV-0.5B achieves a notable jump from 36.41 to 41.94. Although the absolute improvements are moderate, they are meaningful given the benchmark’s difficulty, indicating that \textsc{FOCUS} robustly enhances performance across diverse architectures and scales.

On \text{MuirBench}, which emphasizes multi-image reasoning, \textsc{FOCUS} again provides improvements across nearly all model configurations: InternVL3 family shows an average gain of +0.4\%pts, Qwen2.5-VL-3B achieves +0.93\%pts, and LLaVA-OV family gains +2.14\%pts. These results suggest that \textsc{FOCUS} offers incremental benefits even in complex and ambiguous multi-image settings.

Results on \text{MIRB} demonstrate similar trends. \textsc{FOCUS} generally improves accuracy across model families and sizes, though smaller models such as LLaVA-OV-0.5B show limited or inconsistent gains. The overall upward trend reinforces \textsc{FOCUS} as a model-agnostic enhancement for multi-image reasoning.

\textbf{Results on Video Understanding Benchmark.}
\begin{table}[t]
\centering
\small
\renewcommand{\arraystretch}{1.2}
\setlength{\tabcolsep}{3pt}
\begin{tabular}{l|ccc|ccc}
\toprule
\textbf{Model Size} & \multicolumn{3}{c|}{$<$\textbf{7B}} & \multicolumn{3}{c}{$\geq$\textbf{7B}} \\
\midrule
\diagbox[innerwidth=1.5cm]{\textbf{Family}}{\textbf{Score}} & \textbf{T} & \textbf{V} & \textbf{G} & \textbf{T} & \textbf{V} & \textbf{G} \\
\midrule
Qwen3-VL         & 32.60 & 17.60 & 6.20  & 34.40 & 35.60 & 18.20 \\
\cellcolor{gray!10}+\textsc{FOCUS} & \cellcolor{gray!10}\textbf{33.00} & \cellcolor{gray!10}\textbf{25.80} & \cellcolor{gray!10}\textbf{11.60} & \cellcolor{gray!10}\textbf{35.60} & \cellcolor{gray!10}\textbf{37.00} & \cellcolor{gray!10}\textbf{18.60} \\
\bottomrule
\end{tabular}
\caption{Results on Vinoground with and without \textsc{FOCUS}. Metrics: T (Text), V (Video), G (Group). Bold indicates improvement over the baseline.}
\label{tab:vinoground}
\end{table}
 To examine whether \textsc{FOCUS} generalizes beyond multi-image inputs to video understanding, we evaluate on the Vinoground~\cite{zhang2024vinoground} benchmark.
As shown in Table~\ref{tab:vinoground}, \textsc{FOCUS} consistently improves on Qwen3-VL-4B and Qwen3-VL-8B. These results demonstrate that the benefits of image-wise focused decoding also generalize to the video domain.

\noindent\textbf{Extension to Long-Context Setting.}
We further examine whether \textsc{FOCUS} generalizes to long-context settings~\cite{wang2026mmlongbench}.
Full results and discussion are provided in Appendix~\S\ref{app:longcontext}.

\subsection{Ablation Study}

We conduct ablation studies to assess each component of \textsc{FOCUS}. All experiments are conducted on the Mantis-Eval validation set using Qwen2.5-VL-7B. Unless otherwise specified, we set noise scale $\lambda = 0.3$ and weighting $\alpha = 0.4$.

Table~\ref{tab:ablation_main} shows that removing noise injection passes all images to the model without masking, which prevents image-wise focused inference. The model cannot isolate individual image signals and suffers from cross-image information leakage, leading to a drop in accuracy from 76.19 to 71.43. When noise reference subtraction is disabled, residual artifacts introduced by noise masking remain unsuppressed during contrastive aggregation. This allows spurious visual signals to contaminate the final prediction. As a result, accuracy drops more significantly, from 76.19 to 61.90.

Both components contribute meaningfully to the final performance, confirming that noise injection and contrastive aggregation are complementary and jointly essential to \textsc{FOCUS}.
Appendix~\S\ref{app:more_ablation} reports further ablations on our design choices.
\begin{table}[t]
\centering
\small
\setlength{\tabcolsep}{4pt}
\renewcommand{\arraystretch}{1.2}
\begin{tabular}{l|c|c|c}
\toprule
&  \textbf{w/o Noise ($\lambda$)} &  \textbf{w/o Reference ($\alpha$)} & \textbf{\textsc{FOCUS}} \\
\midrule
Acc. & 71.43 & 61.90 & \textbf{76.19} \\
\bottomrule
\end{tabular}
\caption{Ablation study on each component of \textsc{FOCUS}. We use Qwen2.5-VL-7B on Mantis-Eval validation set.}
\label{tab:ablation_main}
\end{table}

\begin{figure}[t]
  \centering
  \includegraphics[width=\columnwidth]{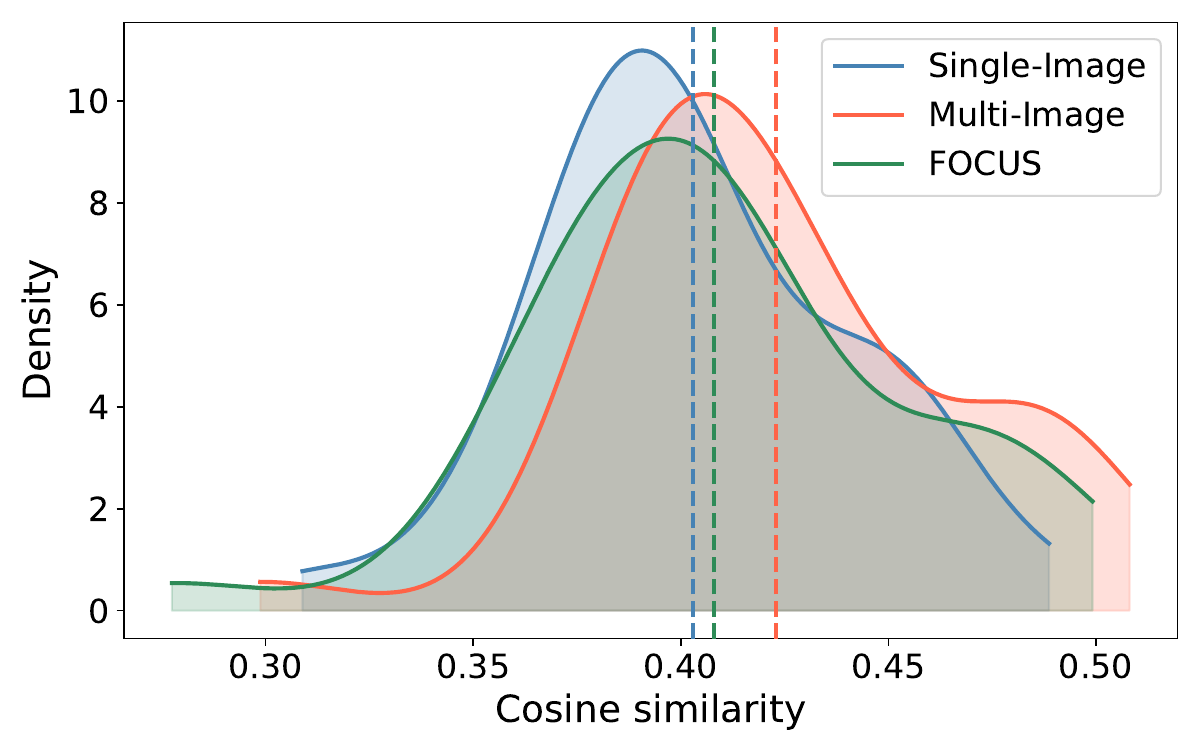}
  \caption{Average pairwise cosine similarity distribution of image 
token representations after the LLM decoder under three conditions. We use Qwen2.5-VL-3B on samples with 4 images each 
from Mantis-Eval.}
  \label{fig:4_rep_sim}
\end{figure}

\subsection{Cross-Image Similarity Analysis}
\label{sec:sim_analysis}
This analysis examines how much visual representations overlap across images in single-image versus multi-image settings, which supports the intuition behind \textsc{FOCUS}.
We extract final hidden states at image token positions from the model across samples from Mantis-Eval. 
For each image, we first average hidden states over its token positions, then compute pairwise cosine similarity across all image combinations per sample.
We compare three conditions:

\begin{itemize}[noitemsep, topsep=2pt, leftmargin=*]
    \item \textit{Single-Image}: Each image is processed independently via a separate forward pass.
    \item \textit{Multi-Image}: All images are processed jointly in a single forward pass.
    \item \textsc{FOCUS}: For each target image, noise is injected into the remaining images and processed jointly.
\end{itemize}

As shown in Figure~\ref{fig:4_rep_sim}, the similarity distribution shifts rightward under joint processing. This suggests that causal attention entangles image representations when multiple images are processed together. \textsc{FOCUS} reduces the mean similarity, nearly recovering the single-image baseline. This indicates that injecting noise into non-target images discourages the model from mixing representations across images, encouraging each image to maintain a more distinct representation during decoding.
We further verify that this pattern persists across larger model scales, including Qwen2.5-VL-7B and Qwen3-VL-8B (provided in Appendix~\S\ref{app:leakage_analysis_scaleup}).

\begin{table}[t]
\centering
\small
\setlength{\tabcolsep}{4pt}
\renewcommand{\arraystretch}{1.2}
\begin{tabular}{c|cc|cc|cc}
\toprule
\multirow{2}{*}{\textbf{Method}} & 
\multicolumn{2}{c|}{\textbf{InternVL3}} & 
\multicolumn{2}{c|}{\textbf{Qwen2.5-VL}} & 
\multicolumn{2}{c}{\textbf{LLaVA-OV}} \\
 & \textbf{2B} & \textbf{8B} & \textbf{3B} & \textbf{7B} & \textbf{0.5B} & \textbf{7B} \\
\midrule
\multicolumn{7}{c}{\textit{Process Independently}} \\
\midrule
Per-Img & 19.61 & 35.71 & 34.57 & 38.18 & 10.53 & 32.82 \\
\midrule
\multicolumn{7}{c}{\textit{Process Jointly}} \\
\midrule
Baseline & 31.32 & \underline{41.86} & 49.27 & 50.53 & \textbf{22.55} & \underline{48.02} \\
Conv-CD & \underline{31.68} & 41.07 & \underline{49.64} & \underline{51.29} & 20.43 & 44.79 \\
SoFA & 30.96 & 40.66 & 49.43 & 51.39 & 12.59 & 42.87 \\
\cellcolor{gray!10}{\textsc{FOCUS}} &   \cellcolor{gray!10}{\textbf{32.36}} &  \cellcolor{gray!10}{\textbf{42.48}} &  \cellcolor{gray!10}{\textbf{50.00}} &  \cellcolor{gray!10}{\textbf{52.30}} &  \cellcolor{gray!10}{\underline{22.13}} &  \cellcolor{gray!10}{\textbf{48.85}} \\
\bottomrule
\end{tabular}
\caption{Comparison of \textsc{FOCUS} with alternative approaches on MIRB. Per-Img: per-image processing. Conv-CD: conventional contrastive decoding. SoFA: \citet{tian2025identifying}.}
\label{tab:single_vcd_ours_comparison}
\end{table}

\begin{figure*}[t]
  \centering
  \includegraphics[width=\textwidth]{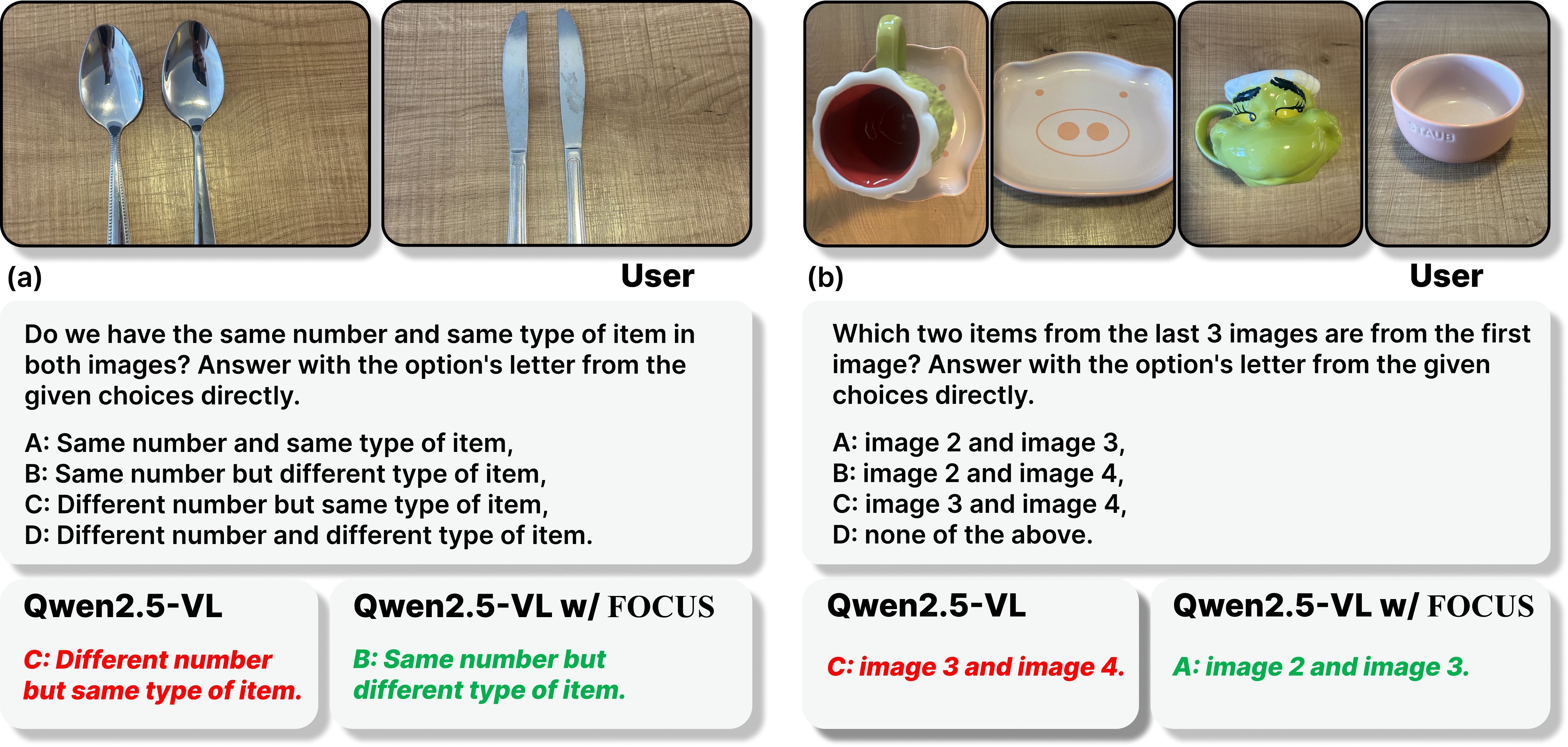}
  \caption{Qualitative results on Mantis-Eval. We compare the Qwen2.5-VL 3B model with and without FOCUS. These samples require the model to interpret each image independently while also integrating information across all images to reach the correct answer. The baseline fails under this setting, however FOCUS preserves per-image understanding and produces the correct responses.}
  \label{fig:3_qualitative}
\end{figure*}

\subsection{Comparison with Alternative Approaches}
\label{sec:Comparison}

We compare \textsc{FOCUS} against three alternative approaches to multi-image reasoning.

\textbf{Per-Image Processing.} 
As discussed in \S\ref{sec:motiv_limitations}, the most straightforward way to avoid cross-image information leakage is to process each image independently. However, this approach inherently sacrifices the model’s ability to perform holistic reasoning across multiple visual inputs, such as leveraging positional or contextual relationships among them. To mitigate this limitation, one possible strategy is to explicitly provide the image order as part of the textual prompt during decoding. 

To examine this idea, we design an experiment where the model generates an independent response for each image. These responses are then concatenated, and the original image order information is added to the prompt to produce a final output. 

As shown in Table~\ref{tab:single_vcd_ours_comparison}, this approach leads to a significant drop in performance compared to the original multi-image setup. Even when the image order is explicitly provided, the model still fails to generate accurate answers. This result suggests that the model effectively utilizes inter-image relationships when multiple images are processed jointly, and that such relationships cannot be easily replaced by simple textual cues.

\textbf{Conventional Contrastive Decoding.}
We further investigate whether contrastive decoding (CD), originally developed for single-image inputs, can be effectively extended to multi-image scenarios. Specifically, we apply VCD~\cite{leng2024mitigating}, a representative CD method for LVLMs.
We expand VCD by contrasting the original multi-image logits with the logits obtained from the same input with added noise.
Unlike our method, which independently masks each image position, VCD contrasts all original inputs with fully noised images. It does not consider cross-image information leakage.

Table~\ref{tab:single_vcd_ours_comparison} shows that conventional CD method in multi-image contexts degrades performance for several model families, even falling below the original model. This demonstrates that simply transferring CD techniques from single-image to multi-image settings not only fails to mitigate information leakage but also further distracts the model's output.

\textbf{SoFA.}
\citet{tian2025identifying} propose SoFA, a training-free method that mitigates position bias of multi-image VL models. As SoFA targets position bias rather than cross-image information leakage, we include it here to examine whether mitigating position bias also alleviates leakage. 

In Table~\ref{tab:single_vcd_ours_comparison}, SoFA does not consistently improve performance either, and similarly drops below the baseline in certain configurations.  This suggests that position bias and cross-image information leakage are distinct problems. In contrast, \textsc{FOCUS} consistently outperforms all alternatives.

\begin{table}[t]
\centering
\small
\renewcommand{\arraystretch}{1.2}
\setlength{\tabcolsep}{3.3pt}
\begin{tabular}{ccccc}
\toprule
\multirow{2}{*}{$N$} & \multicolumn{2}{c}{\textbf{Qwen3-VL-4B}} & \multicolumn{2}{c}{\textbf{Qwen3-VL-8B}} \\
\cmidrule(lr){2-3} \cmidrule(lr){4-5}
 & Baseline & +\textsc{FOCUS} & Baseline & +\textsc{FOCUS} \\
\midrule
2 & 1.38 ± 2.82 & 2.12 ± 2.72 & 1.62 ± 2.91 & 2.20 ± 3.05 \\
3 & 1.00 ± 1.48 & 2.64 ± 1.22 & 1.26 ± 1.78 & 2.67 ± 1.53 \\
4 & 0.43 ± 0.34 & 1.35 ± 1.15 & 1.90 ± 4.26 & 2.47 ± 3.15 \\
5 & 0.43 ± 0.00 & 1.03 ± 0.00 & 0.48 ± 0.00 & 1.04 ± 0.00 \\
\midrule
Acc. & 66.82 & \textbf{68.66} & 65.44 & \textbf{68.66} \\
\bottomrule
\end{tabular}
\caption{Latency (mean $\pm$ std, seconds) and accuracy on Mantis-Eval across 
different numbers of input images.}
\label{tab:acc_tradeoff}
\end{table}
\subsection{Accuracy-Latency Trade Off}
\textsc{FOCUS} introduces moderate latency overhead, which is justified by consistent accuracy gains.
As shown in Table~\ref{tab:acc_tradeoff}, \textsc{FOCUS} yields up to +3.22\%pt accuracy improvement with manageable latency cost. The overhead stays well below the naive $(N{+}1)\times$ cost, as we batch the forward passes into a single inference call. Full details are provided in Appendix~\S\ref{app:implementation_details}.

\subsection{Qualitative Results}
\label{sec:qualitative}
In Figure~\ref{fig:3_qualitative}(a), while the baseline conflates visual content across images, \textsc{FOCUS} correctly differentiates objects and maintains accurate counts (Figure~\ref{fig:3_qualitative}(b)). Additional results are in Appendix~\S\ref{app:additional_qual}.
\section{Conclusion}
In this work, we empirically characterize an underexplored limitation of LVLMs in multi-image settings: \textit{cross-image information leakage}, where models tend to mix visual information across images rather than reasoning over each independently. To address this, we propose \textsc{FOCUS}, a training-free inference method that masks non-target images with noise to guide the model to attend to one image at a time, and aggregates the resulting logits through contrastive logit aggregation. We validate \textsc{FOCUS} across diverse LVLM families and benchmarks, and demonstrate that its benefits extend to video understanding. Our results show that \textsc{FOCUS} offers a practical and generalizable inference-time solution for more reliable multi-image reasoning.
\section*{Limitations}

\noindent\textbf{Baseline Reproducibility.}
Since LVLM performance is sensitive to generation configuration and instruction prompts, the baseline results reported in this work may differ from those in the official model reports of \citet{bai2025qwen2, bai2025qwen3, Chen_2024_CVPR, 
li2024allava}.

\noindent\textbf{Hyperparameter Sensitivity.}
The hyperparameters $\lambda$ and $\alpha$ are selected per model and benchmark via a validation set, and their generalization to unseen settings without validation data has not been verified. Full hyperparameter tables are provided in the Appendix~\S\ref{app:implementation_details}.

\noindent\textbf{Inference Overhead.}
\textsc{FOCUS} requires $N{+}1$ forward passes for $N$ input images, introducing additional inference cost. In practice, this overhead is substantially smaller than the naive $(N{+}1)\times$ cost, since the passes share the same sequence length and are executed as a single batched call, yielding a 2.9 to 3.8$\times$ speedup over sequential execution (Appendix~\S\ref{app:inference_cost}). Although the cost grows with the number of images, the batched implementation keeps it well below the naive $(N{+}1)\times$ bound, making \textsc{FOCUS} practical in typical multi-image settings.

\noindent\textbf{Long-Context Applicability.}
Long-context multi-image understanding is another large body of research with distinct objectives from our work, which focuses on disambiguating visual content across a small set of images. While \textsc{FOCUS} shows promising gains on long-context tasks requiring cross-image comparison, it yields mixed results on retrieval-heavy tasks where the bottleneck lies in locating a single relevant image rather than disambiguating across multiple images.
\section*{Acknowledgements}

This work was partly supported by the Institute of Information \& Communications Technology Planning \& Evaluation (IITP) grants funded by the Korea government (MSIT) (No. RS-2025-25441313, Professional AI Talent Development Program for Multimodal AI Agents, Contribution: 50\%; No. IITP-RS-2026-25547954, Artificial Intelligence Innovation Human Resources Development, Contribution: 10\%) and the National Research Foundation of Korea (NRF) grant funded by the Korea government (MSIT) (No. RS-2024-00350430, Contribution: 40\%). This work was also supported by the Hyundai Motor Chung Mong-Koo Foundation.


\bibliography{acl/custom}

\appendix\clearpage
\setcounter{page}{1}
\appendix
\setcounter{section}{0}
\renewcommand{\thesection}{\Alph{section}}
\renewcommand{\thesubsection}{\Alph{section}.\arabic{subsection}}

\renewcommand{\thefigure}{\Alph{section}\arabic{figure}}
\renewcommand{\thetable}{\Alph{section}\arabic{table}}

\makeatletter
\@addtoreset{figure}{section}
\@addtoreset{table}{section}
\makeatother

\setcounter{figure}{0}
\setcounter{table}{0}

\section{Related Work}
While LVLMs acheive strong performance on standard single-image tasks~\cite{Chen_2024_CVPR, li2024allava}, their ability to understand multi-image remains underdeveloped.
Many open-source LVLMs~\cite{dai2023instructblip, liu2023visual, zhu2024minigpt, liu2024llavanext} are trained under the assumption of single-image inputs, and therefore struggle to generalize when presented with multiple images simultaneously~\cite{wang2025muirbench}.

\noindent\textbf{Multi-Image LVLMs.} To address this gap, several recent LVLMs~\cite{awadalla2023openflamingo, sun2024generative, laurenccon2023obelics, lin2024vila, li2024bllava} have been trained using large-scale interleaved image-text datasets~\cite{laurenccon2023obelics, li2024bllava}.
For example, \citet{laurenccon2023obelics} comprises hundreds of millions of interleaved image-text data that support multiple, contextually related images.
While these approaches improve multi-image understanding capabilities, they require extensive training resources~\cite{jiang2024mantis} and suffer from limited reusability and flexibility due to their reliance on model-specific architectures.
However, even recent state-of-the-art models continue to show notable performance gaps in multi-image settings~\cite{wang2025muirbench}, and the underlying causes remain insufficiently studied. 
\citet{tian2025identifying} identify position bias in multi-image settings, where predictions vary with image order. In contrast, our work targets cross-image information leakage, where the model confuses visual content across images regardless of their order. 

\noindent\textbf{Training-free Methods.} 
Training-free methods guide generation without additional training or architectural changes. These approaches manipulate logits or hidden states during inference to steer outputs toward desired properties~\cite{li-etal-2023-contrastive, malkin-etal-2022-coherence, shi-etal-2024-trusting}.
Within LVLMs, a growing body of work~\cite{leng2024mitigating,huang2024opera,chen2024halc,wang2025mllm, park2025convis,Chen_2025_CVPR} has shown promising results in reducing hallucination, but they have predominantly been developed under single-image assumptions~\cite{Suo_2025_CVPR, dong2025inter, suo2025octopus, an2025mitigating}.
\citet{tian2025identifying} propose a training-free method that mitigates position bias by modifying inter-image attention masks. However, their method targets position bias rather than cross-image information leakage. In contrast, our method directly addresses cross-image information leakage through logit manipulation.

\section{Benchmarks}
\label{app:benchmarks}

\noindent\textbf{Winoground~\cite{thrush2022winoground}} evaluates whether a model can correctly associate captions with images or vice versa. It contains 400 samples, each consisting of two images and two captions, $\{(I_1, T_1), (I_2, T_2)\}$, with subtle differences. The two captions contain the same words but in a different order. The model must assign the correct caption to each image.

\noindent\textbf{VisMin~\cite{awal2024vismin}} builds on this setup by focusing on 2K image-text pairs with minimal semantic differences. It is organized into four semantic categories, targeting more fine-grained discrimination.
We further split these into VisMin-Easy and VisMin-Hard, where VisMin-Easy contains Object and Attribute categories, and VisMin-Hard contains Relation and Counting categories. 

\noindent\textbf{VisMin-Easy} tests mutually exclusive properties by nature, as an object cannot simultaneously belong to two distinct categories, nor can an attribute hold two contradictory values at once.

\noindent\textbf{VisMin-Hard} tests more inherently ambiguous categories, as spatial relations can co-occur across images and numerical counts are susceptible to cross-image information leakage.

\noindent\textbf{Evaluation metrics for Winoground and VisMin.}
We adopt the generative LVLM evaluation protocol introduced in \citet{awal2024vismin}, which is designed to assess open-ended generation by three metrics.
\begin{itemize}[noitemsep, topsep=2pt, leftmargin=*]
    \item \textit{Text Score}: For each input set $\{I_1, T_1, T_2\}$ and $\{I_2, T_1, T_2\}$, the model is prompted to choose the caption that best matches the image. A point is awarded only if the model selects the correct caption for both inputs. This metric evaluates single-image caption grounding. \item \textit{Image Score}: For each input set $\{I_1, I_2, T_1\}$ and $\{I_1, I_2, T_2\}$, the model is prompted to select the image that best matches the caption. A point is awarded only if the correct image is chosen for both inputs. This metric evaluates comparative reasoning and primarily reflects multi-image understanding.
    \item \textit{Group Score}: It requires both the Text Score and the Image Score to be correct. It reflects the model’s ability to reason consistently across both single-image and multi-image contexts.
\end{itemize}

\noindent\textbf{Recet Multi-image Benchmarks.} We also evaluate on three recent multi-image benchmarks for LVLMs.
Mantis-Eval~\cite{jiang2024mantis} covers diverse topics such as size perception and weight comparison. Each sample is crafted by annotators to require deep cross-image understanding.
MuirBench~\cite{wang2025muirbench} assesses 12 multi-image reasoning skills across 10 inter-image relation types. It pairs each instance with an unanswerable variant to test fine-grained discrimination.
MIRB~\cite{zhao2024benchmarking} evaluates cross-image comparison and reasoning across four categories: perception, visual knowledge, reasoning, and multi-hop reasoning.
All three benchmarks are evaluated by accuracy.

\noindent\textbf{Vinoground}~\cite{zhang2024vinoground} is a temporal counterfactual benchmark consisting of 1,000 short video-caption pairs designed to evaluate whether models can distinguish temporal differences between actions and object transformations. Models are evaluated with Text, Video, and Group scores, following the same protocol as Winoground.

\noindent\textbf{MMLongBench}~\cite{wang2026mmlongbench} is a long-context vision-language benchmark consisting of 13,331 examples spanning five task categories, including Visual RAG and Many-Shot ICL, evaluated at standardized input lengths from 8K to 128K tokens. We evaluate on four categories: ICL, DocVQA, VRAG, and NIAH, using Qwen2.5-VL-7B at 8K context length.

\subsection{Data Usage and Licensing}
All datasets used in this work are employed strictly for research and evaluation purposes. Winoground~\cite{thrush2022winoground} is used under the Meta Images Research License, which permits use for non-commercial research and evaluation. 
Mantis-Eval~\cite{jiang2024mantis} is used under the Apache 2.0 License. 
MuirBench~\cite{wang2025muirbench} and MIRB~\cite{zhao2024benchmarking} are used in accordance with their respective terms of use, which permit research use with appropriate attribution. All datasets are used solely for evaluation without any modification or redistribution.

\section{Implementation Details}
\label{app:implementation_details}

\noindent\textbf{Models.}
We evaluate on four representative LVLM families: InternVL3 (2B, 8B)~\cite{Chen_2024_CVPR}, Qwen2.5-VL (3B, 7B)~\cite{bai2025qwen2}, Qwen3-VL (4B, 8B, 32B, 30B-A3B)~\cite{bai2025qwen3}, and LLaVA-OneVision (0.5B, 7B)~\cite{li2024allava}. All models are evaluated in a frozen state without any fine-tuning; only the decoding phase is modified. Model checkpoints are publicly 
available on Hugging Face.\footnote{\url{https://huggingface.co/OpenGVLab/InternVL3-2B}, 
\url{https://huggingface.co/OpenGVLab/InternVL3-8B}, 
\url{https://huggingface.co/Qwen/Qwen2.5-VL-3B-Instruct}, 
\url{https://huggingface.co/Qwen/Qwen2.5-VL-7B-Instruct}, 
\url{https://huggingface.co/Qwen/Qwen3-VL-4B-Instruct}, 
\url{https://huggingface.co/Qwen/Qwen3-VL-8B-Instruct},
\url{https://huggingface.co/Qwen/Qwen3-VL-32B-Instruct}
\url{https://huggingface.co/Qwen/Qwen3-VL-30B-A3B-Instruct},
\url{https://huggingface.co/llava-hf/llava-onevision-qwen2-0.5b-ov-hf},
\url{https://huggingface.co/llava-hf/llava-onevision-qwen2-7b-ov-hf}}

\noindent\textbf{Decoding Setup.}
We use multinomial sampling with temperature $T=0.2$.

\noindent\textbf{Noise Masking.}
Each image $v_i$ is masked using additive uniform noise as $v'_i = (1 - \lambda) \cdot v_i + \lambda \cdot \mathcal{U}(0, 1)$, where $\lambda \in [0, 1]$ is a hyperparameter controlling the degree of corruption. The masking is applied in the normalized image space using ImageNet statistics ($\mu = [0.485, 0.456, 0.406]$, $\sigma = [0.229, 0.224, 0.225]$), then denormalized back to pixel space and clipped to $[0, 1]$. The optimal value of $\lambda$ is selected via validation per model and benchmark.

\noindent\textbf{Hyperparameter Tuning.}
All hyperparameters including noise type, noise scale $\lambda$, and aggregation weight $\alpha$ are selected per model and benchmark via validation. For VisMin, we randomly sample a subset of the training set to construct a validation split. For benchmarks without official validation sets, we randomly choose 10\% of the test data for this purpose.

\begin{table}[t]
\centering
\small
\setlength{\tabcolsep}{3pt}
\begin{tabular}{l c cc cc cc}
\toprule
\multirow{2}{*}{\textbf{Dataset}} &
\multirow{2}{*}{\textbf{Var.}} &
\multicolumn{2}{c}{\textbf{InternVL3}} &
\multicolumn{2}{c}{\textbf{Qwen2.5-VL}} &
\multicolumn{2}{c}{\textbf{LLaVA-OV}} \\
\cmidrule(lr){3-4} \cmidrule(lr){5-6} \cmidrule(lr){7-8}
 & & \textbf{2B} & \textbf{8B} & \textbf{3B} & \textbf{7B} & \textbf{0.5B} & \textbf{7B} \\
\midrule
\multirow{2}{*}{Winog.}
& $\lambda$ & 1 & 1 & 0.6 & 1 & 1 & 1 \\
& $\alpha$  & 0.6 & 0.8 & 0.15 & 0.2 & 0.1 & 0.3 \\
\midrule
\multirow{2}{*}{VisMin}
& $\lambda$ & 1 & 0.7 & 0.5 & 0.7 & 1 & 0.9 \\
& $\alpha$  & 0.4 & 0.7 & 0.5 & 0.15 & 0.1 & 0.2 \\
\midrule
\multirow{2}{*}{Mantis}
& $\lambda$ & 0.8 & 0.3 & 0.8 & 0.3 & 0.1 & 0.2 \\
& $\alpha$  & 0.1 & 0.1 & 0.5 & 0.4 & 0.2 & 0.5 \\
\midrule
\multirow{2}{*}{Muir.}
& $\lambda$ & 0.2 & 0.5 & 0.8 & 0.6 & 0.1 & 1 \\
& $\alpha$  & 0.1 & 0.1 & 0.1 & 0.1 & 0.1 & 0.1 \\
\midrule
\multirow{2}{*}{MIRB}
& $\lambda$ & 0.3 & 0.2 & 0.1 & 0.3 & 0.1 & 0.1 \\
& $\alpha$  & 0.1 & -0.15 & -0.2 & 0.1 & -0.15 & 0.1 \\
\bottomrule
\end{tabular}
\caption{Hyperparameters used for evaluation. Search space: $\lambda \in (0,1)$ and $\alpha \in (-1,1)$.}
\label{tab:hyperparameter}
\end{table}

\noindent\textbf{Hyperparameter Settings} To ensure reproducibility, we provide the list of hyperparameters used in our experiments in Table~\ref{tab:hyperparameter}. The two primary hyperparameters are the noise scale $\lambda$ and the contrastive weight $\alpha$. Intuitively, $\lambda$ controls the strength of visual noise injection, while $\alpha$ balances the contrastive aggregation weight.
We perform a random search over $\lambda \in (0, 1)$ and $\alpha \in (-1, 1)$ using validation set. Across datasets and model sizes, we observe two consistent trends: (i) the noise scale $\lambda$ generally converges to relatively large values, and (ii) the contrastive weight $\alpha$ remains stable around a small magnitude. These patterns hold regardless of the underlying architecture or parameter count. More specifically, we observe that $\lambda$ tends to saturate at 1.0 for most cases in the Winoground dataset, while $\alpha$ is consistently selected as 0.1 for MuirBench.

\noindent\textbf{More Details.}
Experiments are conducted on NVIDIA A100 (80GB) and RTX 6000 Pro GPUs. For 
memory efficiency, images are resized to $512 \times 512$ for MuirBench and 
MIRB evaluations.

\noindent\textbf{SoFA Reproduction.}
As the official implementation of SoFA~\cite{tian2025identifying} is unavailable, we reproduce it based on the original paper. We set $\sigma = 0.5$ and apply the interpolation every two layers, following the experimental setup described in the original work.

\noindent\textbf{Accuracy-Latency Trade-off Details.}
We conduct latency experiments on Mantis-Eval~\cite{jiang2024mantis} using Qwen3-VL-4B and Qwen3-VL-8B across $N \in \{2, 3, 4, 5\}$ input images, reporting wall-clock end-to-end latency per sample. Since Mantis-Eval has an uneven distribution of samples across image counts, where the $N=4$ and $N=5$ subsets contain significantly fewer samples than $N=2$ and $N=3$, we report average latency per $N$ and average accuracy over the full test split separately. For the same reason, latency should be compared between the baseline and \textsc{FOCUS} within each $N$ rather than across different $N$, as each subset has its own answer-length distribution.

\noindent\textbf{Batched Implementation.}
Rather than running the $N{+}1$ forward passes sequentially, we execute them as a single batched inference call. All $N{+}1$ passes share the same sequence length, so batching is efficient on the LLM side. This keeps the practical overhead far below the naive $(N{+}1)\times$ cost.

\noindent\textbf{Inference Cost}
\label{app:inference_cost}
We run a controlled synthetic experiment to isolate how the number of input images affects latency. For each $N$, we generate a fixed set of $N$ random image tensors of shape $(512, 512, 3)$ and repeat each run three times. We fix the maximum number of new tokens to 64. We report the average per-sample latency (s/sample).
\begin{table}[t]
\centering
\small
\setlength{\tabcolsep}{6pt}
\renewcommand{\arraystretch}{1.2}
\begin{tabular}{c|cc|c}
\toprule
 & \multicolumn{2}{c|}{\textbf{Latency}} & \textbf{Speedup} \\
\textbf{$N$} & \textbf{Baseline} & \textbf{\textsc{FOCUS}} & \textbf{vs. $(N{+}1)\times$} \\
\midrule
2  & 1.87 & 1.96  & 2.86 \\
4  & 1.93 & 2.58  & 3.73 \\
8  & 2.11 & 4.97  & 3.81 \\
16 & 2.47 & 14.26 & 2.94 \\
\bottomrule
\end{tabular}
\caption{Inference cost of \textsc{FOCUS} versus the number of input images $N$. Per-sample latency in seconds, measured on synthetic $(512, 512, 3)$ tensors and averaged over three runs. The last column is the speedup of our batched implementation over the naive $(N{+}1)\times$ cost.}
\label{tab:inference_cost}
\end{table}
As shown in Table~\ref{tab:inference_cost}, batched inference yields a 2.9$\times$ to 3.8$\times$ speedup over the naive $(N{+}1)\times$ cost across all tested image counts. The overhead is modest in the small-$N$, and our batched implementation keeps it well below the naive $(N{+}1)\times$ cost as $N$ grows.
\section{Additional Experiments}
\subsection{Cross-Image Information Leakage in Larger Models}
\label{app:leakage_analysis_scaleup}
\begin{figure*}[t]
\centering
  \includegraphics[width=\textwidth]{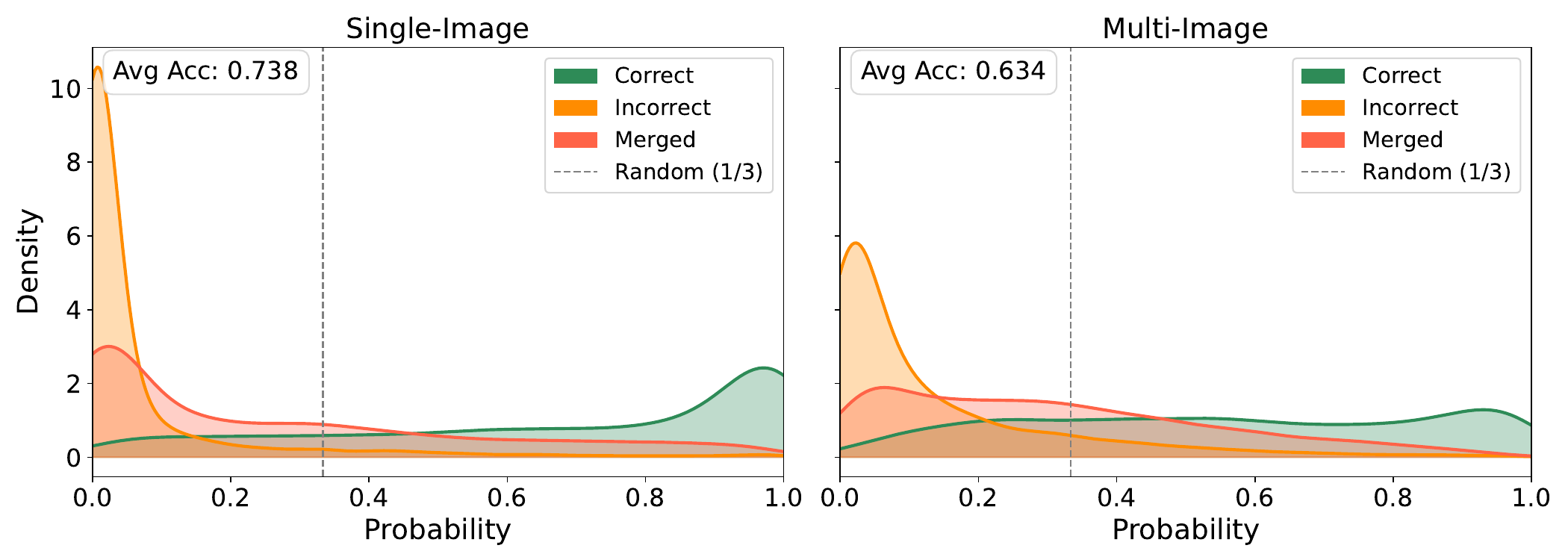}
  \vspace{0.5em}
  \includegraphics[width=\textwidth]{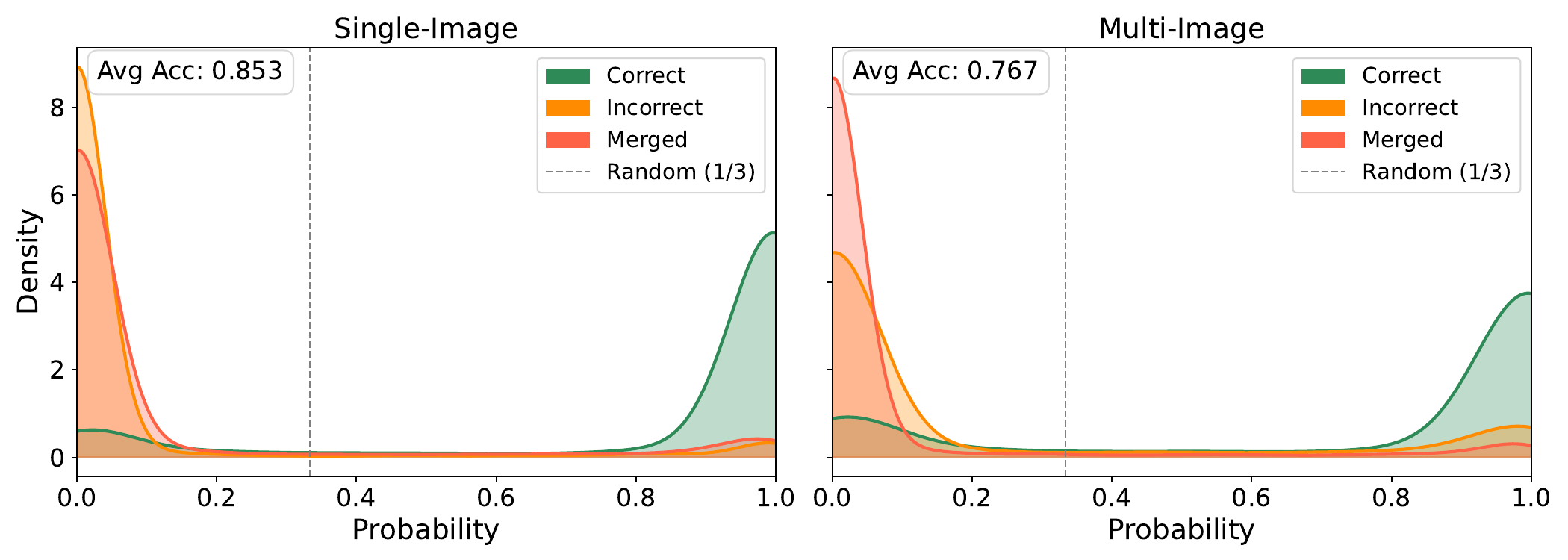}
  \caption{Estimation of probability density function of model predictions over three answer options, replicating the analysis of Figure~\ref{fig:motivation_analysis} at larger model scales: 
  Qwen2.5-VL-7B (top) and Qwen3-VL-8B (bottom). Accuracy consistently degrades under multi-image conditions across both scales, confirming that cross-image information leakage is not an artifact of smaller models.}
  \label{fig:A_motivation_scaleup}
\end{figure*}
\begin{figure*}[t]
\centering
  \includegraphics[width=0.48\linewidth]{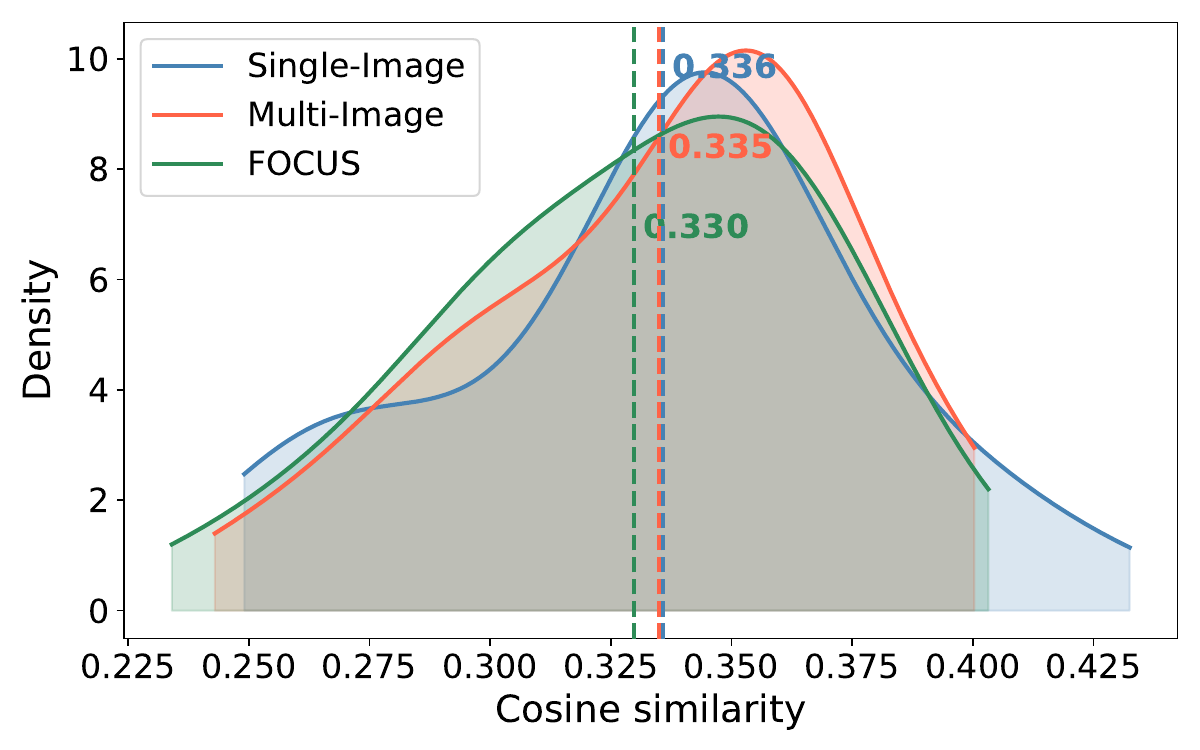} \hfill
  \includegraphics[width=0.48\linewidth]{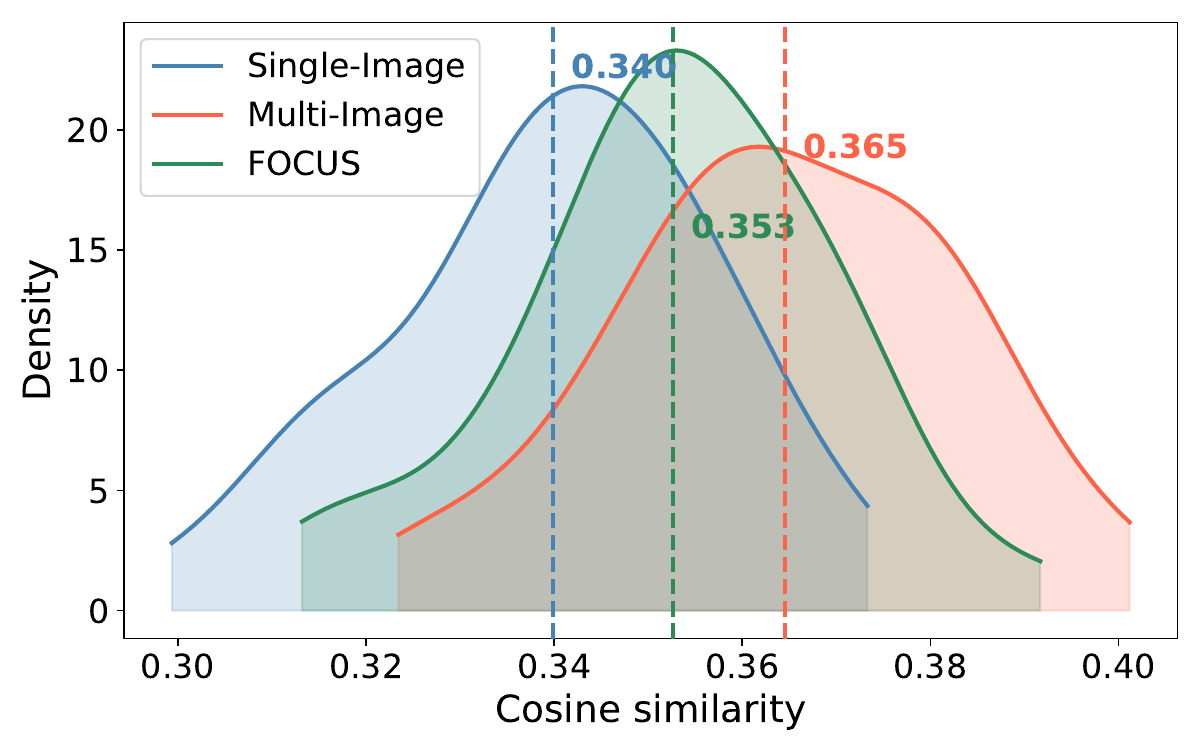}
  \caption{Average pairwise cosine similarity distribution of image token representations under three conditions (single-image, multi-image, and +\textsc{FOCUS}), replicating the analysis of Figure~\ref{fig:4_rep_sim} at larger model scales: Qwen2.5-VL-7B (left) and Qwen3-VL-8B (right).}
  \label{fig:A_sim_scaleup}
\end{figure*}

We replicate the analyses from Section~\S\ref{sec:leakage_motiv} and Section~\S\ref{sec:sim_analysis} on Qwen2.5-VL-7B and Qwen3-VL-8B to verify that cross-image information leakage is not an artifact of smaller models.

\noindent\textbf{Behavioral Analysis.}
Following the same setup as Section~\ref{sec:leakage_motiv}, the model selects among a correct caption, an incorrect caption from another image, and a merged caption combining both. As shown in Figure~\ref{fig:A_motivation_scaleup}, accuracy consistently degrades under multi-image conditions across both scales. For Qwen2.5-VL-7B, merged-caption selection increases under multi-image input, 
a direct behavioral signature of cross-image confusion. For Qwen3-VL-8B, merged-caption selection decreases, yet accuracy still drops substantially (85.30 → 76.66). This suggests that stronger models may exhibit leakage.

\noindent\textbf{Representational Analysis.}
As shown in Figure~\ref{fig:A_sim_scaleup}, for Qwen3-VL-8B, multi-image input increases hidden state similarity (0.340 → 0.366), and \textsc{FOCUS} reduces it back toward the single-image baseline (0.353), consistent with cross-image entanglement at the representational level. For Qwen2.5-VL-7B, the difference 
between single- and multi-image conditions is minimal (0.336 vs. 0.335). Nevertheless, \textsc{FOCUS} further reduces similarity (0.330), suggesting active suppression of cross-image leakage even when the effect is not fully captured by this metric. These results indicate that cross-image information leakage persists across model scales, though the degree and type of degradation vary with model capacity.

\subsection{Results on Larger Models}
\label{app:moe}
\begin{table}[t]
\centering
\small
\renewcommand{\arraystretch}{1.2}
\begin{tabular}{llccc}
\toprule
\textbf{Benchmark} & \textbf{Metric} & \textbf{Baseline} & \textbf{+\textsc{FOCUS}} \\
\midrule
\multirow{3}{*}{VisMin} 
    & T & 91.06 & 91.06 \\
    & I & 70.12 & \textbf{73.51} \\
    & G & 67.88 & \textbf{70.98} \\
\midrule
Mantis-Eval & Acc. & 64.06 & \textbf{66.36} \\
\bottomrule
\end{tabular}
\caption{Results on Qwen3-VL-30B-A3B (MoE) on VisMin and Mantis-Eval. 
\textsc{FOCUS} consistently improves Image Score (+3.39 pp), Group Score 
(+3.10 pp), and Mantis-Eval accuracy (+2.30 pp), while Text Score remains 
unchanged.}
\label{tab:moe}
\end{table}
\begin{table}[t]
\centering
\small
\setlength{\tabcolsep}{6pt}
\renewcommand{\arraystretch}{1.2}
\begin{tabular}{llcc}
\toprule
\textbf{Benchmark} & \textbf{Metric} & \textbf{Baseline} & \textbf{+\textsc{FOCUS}} \\
\midrule
\multirow{3}{*}{Winoground}
 & T & 81.50 & \textbf{82.50} \\
 & I & 54.75 & \textbf{70.50} \\
 & G & 49.75 & \textbf{65.50} \\
\midrule
\multirow{3}{*}{VisMin}
 & T & \textbf{91.42} & \textbf{91.42} \\
 & I & 88.08 & \textbf{90.12} \\
 & G & 83.89 & \textbf{85.47} \\
\midrule
Mantis-Eval & Acc. & 74.65 & \textbf{75.58} \\
\bottomrule
\end{tabular}
\caption{Results with Qwen3-VL-32B. T, I, and G denote Text, Image, and Group score. VisMin results are averaged over the Object, Attribute, and Counting categories.}
\label{tab:qwen3vl32b}
\end{table}
To verify that \textsc{FOCUS} generalizes beyond the model scales and architectures in our main experiments, we evaluate two additional models. We first evaluate Qwen3-VL-30B-A3B, a MoE architecture. As shown in Table~\ref{tab:moe}, \textsc{FOCUS} consistently improves multi-image relevant metrics across both benchmarks, confirming that its benefits are not limited to dense architectures.

We further evaluate Qwen3-VL-32B on Winoground, VisMin, and Mantis-Eval. As shown in Table~\ref{tab:qwen3vl32b}, \textsc{FOCUS} yields substantial gains on Winoground, improving the Image Score from 54.75 to 70.50 and the Group Score from 49.75 to 65.50. These results indicate that \textsc{FOCUS} remains effective at larger scale, and that its gains are not an artifact of smaller architectures.

\subsection{Extension to Long-Context Setting.}
\label{app:longcontext}
\begin{table}[ht]
\centering
\small
\renewcommand{\arraystretch}{1.2}
\setlength{\tabcolsep}{1.0pt}
\begin{tabular}{llccc}
\toprule
\textbf{Category} & \textbf{Dataset} & \textbf{Metric} & \textbf{Baseline} & +\textbf{\textsc{FOCUS}} \\
\midrule
\multirow{3}{*}{ICL} 
    & Cars        & Acc      & 20.75 & \textbf{26.42} \\
    & Food        & Acc      & 24.00 & \textbf{34.00} \\
    & iNat        & Acc      & 18.00 & \textbf{22.00} \\
\midrule
\multirow{3}{*}{DocVQA} 
    & LongDocURL  & doc\_qa  & 58.94 & \textbf{60.60} \\
    & MMLongDoc   & doc\_qa  & 51.05 & \textbf{53.29} \\
    & SlideVQA    & doc\_qa  & 65.75 & \textbf{68.68} \\
\midrule
\multirow{2}{*}{VRAG} 
    & InfoSeek    & SubEM    & \textbf{60.76} & 59.78 \\
    & ViQuAE      & SubEM    & \textbf{55.68} & 54.19 \\
\midrule
\multirow{5}{*}{NIAH} 
    & VH-Single          & Acc       & 55.57 & \textbf{55.85} \\
    & VH-Multi           & Acc       & 54.37 & \textbf{54.77} \\
    & MM-NIAH Count      & Soft Acc. & 36.88 & \textbf{38.95} \\
    & MM-NIAH Retriev.  & Acc.      & \textbf{25.56} & 25.06 \\
    & MM-NIAH Reason.  & Acc.      & \textbf{51.87} & 51.52 \\
\midrule
\multicolumn{3}{l}{\textbf{Average}} & 44.55 & \textbf{46.54} \\
\bottomrule
\end{tabular}
\caption{Results on MMLongBench~\cite{wang2026mmlongbench} using Qwen2.5-VL-7B. 
\textsc{FOCUS} shows consistent gains on ICL and DocVQA tasks requiring cross-image comparison.}
\label{tab:mmlongbench}
\end{table}
We examine whether \textsc{FOCUS} generalizes to long-context settings by evaluating on MMLongBench~\cite{wang2026mmlongbench} across four task categories (ICL, DocVQA, VRAG, and NIAH) using Qwen2.5-VL-7B at 8K context length. Hyperparameters are set to $\alpha=0.1$ and $\lambda=0.1$ for most datasets, with per-dataset selection for a subset of tasks. Since long-context multi-image understanding is a large body of research~\cite{wang2024needle} with distinct objectives from our work, we do not claim full applicability to this setting. Nevertheless, as shown in Table~\ref{tab:mmlongbench}, \textsc{FOCUS} shows promising gains on tasks that align with our design objective of cross-image disambiguation, particularly on ICL and DocVQA tasks. Results are mixed on VRAG and NIAH tasks, where the bottleneck lies in locating a single relevant image within a long context rather than disambiguating across multiple images simultaneously.

\subsection{Ablation Study}
\label{app:more_ablation}

We conduct ablation studies to assess each component in FOCUS. All experiments are conducted on the Mantis validation set using Qwen2.5-VL 7B. If not specified, we set noise scale $\lambda = 0.3$ and weighting $\alpha = 0.4$.

\textbf{Effect of Noise Type.}
In Table~\ref{tab:ablation_combined} (a), we compare three types of noise: Gaussian, Impulse, and Uniform for masking non-target images during FOCUS inference. Although all noise variants serve to suppress irrelevant image cues, their effectiveness varies. Uniform noise yields the highest accuracy at 76.19\%. In contrast, Gaussian and Impulse noise achieves comparatively lower accuracy.

\textbf{Effect of $\lambda$ (Noise Scale).}
Table~\ref{tab:ablation_combined} (b) reports the effect of the noise strength $\lambda$, which controls how much noise is applied to non-target images. A low value $\lambda = 0.1$ moderately improves performance to 71.43\%, but is insufficient to fully suppress leakage. A high value $\lambda = 1.0$ degrades performance to 52.38\% by overly corrupting the image. Our default setting $\lambda = 0.3$ yields the highest performance at 76.19\%, demonstrating that balancing semantic masking and structural retention is key.

\textbf{Effect of $\alpha$ (Weight).}
Table~\ref{tab:ablation_combined} (c) shows how the contrastive weight $\alpha$ used to subtract noise reference $f_\text{noise}$ affects performance. A small $\alpha = 0.1$ achieves 66.67\%, indicating a benefit from incorporating the noise-only reference. Increasing $\alpha$ to 0.4 (our choice) leads to the best result at 76.19\%. However, setting $\alpha = 1.0$ reduces performance to 61.90\%, likely due to over-suppressing the clean image logits. This suggests that contrastive suppression should be applied carefully to preserve essential signal while reducing information leakage.
\begin{table}[t]
\centering
\small
\setlength{\tabcolsep}{8pt}
\renewcommand{\arraystretch}{1.2}
\begin{tabular}{lc|lc|lc}
\toprule
\multicolumn{2}{c|}{\textbf{(a) Noise Type}} & 
\multicolumn{2}{c|}{\textbf{(b) Noise Scale}} & 
\multicolumn{2}{c}{\textbf{(c) Weight}} \\
\textbf{Variant} & \textbf{Acc.} & 
\textbf{$\lambda$} & \textbf{Acc.} & 
\textbf{$\alpha$} & \textbf{Acc.} \\
\midrule
Gaussian & 71.43 & 0.1 & 71.43 & 0.1 & 66.67 \\
Impulse  & 66.67 & \textbf{0.3} & \textbf{76.19} & \textbf{0.4} & \textbf{76.19} \\
Uniform  & \textbf{76.19} & 1.0 & 52.38 & 1.0 & 61.90 \\
\bottomrule
\end{tabular}
\caption{Impact of \textsc{FOCUS} design choices: noise type, masking strength $\lambda$, and contrastive weight $\alpha$. Mantis validation accuracies using Qwen2.5-VL-7B. Bold indicates best per group.}
\label{tab:ablation_combined}
\end{table}

\textbf{Effect of Reference Type}
Table~\ref{tab:ablation_reference} compares three choices of reference type on the Mantis-Eval test set. We hold the per-image noise term fixed and vary only the reference. is designed to cancel the spurious signal that masking introduces (i.e., the added noise), not genuine image content. A clean image reference removes real visual information along with the artifact. It reduces accuracy to 67.28\%, worse than using no reference at all (69.12\%). Pure noise instead improves over that baseline, reaching 70.05\%. This confirms that the clean-image reference removes useful visual evidence, while the noise reference cancels only the masking artifact.
\begin{table}[t]
\centering
\small
\setlength{\tabcolsep}{8pt}
\renewcommand{\arraystretch}{1.2}
\begin{tabular}{l|c}
\toprule
\textbf{Reference Type} & \textbf{Mantis-Eval (Test)} \\
\midrule
w/o Reference & 69.12 \\
Clean Image as Reference & 67.28 \\
\textsc{FOCUS} (Noise Reference) & \textbf{70.05} \\
\bottomrule
\end{tabular}
\caption{Ablation on the choice of reference type, using Qwen2.5-VL-7B on the Mantis-Eval test set. The per-image noise term is held fixed and only the reference type varies.}
\label{tab:ablation_reference}
\end{table}

\subsection{Detailed VisMin Results}

\begin{table*}[t]
\centering
\small
\setlength{\tabcolsep}{2pt}
\renewcommand{\arraystretch}{1.2}
\begin{tabularx}{\textwidth}{l *{12}{>{\centering\arraybackslash}X}}
\toprule
\textbf{Category} &
  \multicolumn{3}{c}{\textbf{Object}} &
  \multicolumn{3}{c}{\textbf{Attribute}} &
  \multicolumn{3}{c}{\textbf{Relation}} &
  \multicolumn{3}{c}{\textbf{Counting}} \\
\cmidrule(lr){2-4} \cmidrule(lr){5-7} \cmidrule(lr){8-10} \cmidrule(lr){11-13}
& \textbf{T} & \textbf{I} & \textbf{G} 
& \textbf{T} & \textbf{I} & \textbf{G}
& \textbf{T} & \textbf{I} & \textbf{G}
& \textbf{T} & \textbf{I} & \textbf{G} \\
\midrule
InternVL3-2B & 95.68 & 36.10 & 35.92 & 90.82 & 23.81 & 22.79 & \textbf{60.77} & 1.29 & 0.96 & \textbf{70.29} & 12.39 & 11.38 \\
\rowcolor{gray!10}\ \ + \textsc{FOCUS} & \textbf{95.85} & \underline{\textbf{78.58}} & \underline{\textbf{76.86}} & 90.82 & \underline{\textbf{70.07}} & \underline{\textbf{65.65}} & 60.13 & \textbf{10.93} & \textbf{8.36} & 69.61 & \underline{\textbf{42.61}} & \underline{\textbf{39.56}} \\
InternVL3-8B & 96.03 & \textbf{86.01} & \textbf{83.94} & 93.20 & \textbf{83.67} & \textbf{79.25} & 90.68 & 47.43 & 46.14 & 75.04 & 47.88 & 43.46 \\
\rowcolor{gray!10}\ \ + \textsc{FOCUS} & \textbf{95.85} & 85.49 & 83.77 & \underline{\textbf{93.54}} & 80.61 & 77.89 & 89.87 & \textbf{50.64} & \textbf{48.71} & \textbf{75.21} & \textbf{54.67} & \textbf{49.07} \\
QwenVL2.5-3B & \textbf{94.82} & 48.19 & 47.84 & 89.46 & 47.28 & 44.56 & 83.44 & 24.12 & 20.90 & 71.31 & 30.05 & 24.96 \\
\rowcolor{gray!10}\ \ + \textsc{FOCUS} & 94.65 & \textbf{60.10} & \textbf{59.93} & 89.46 & \textbf{53.74} & \textbf{51.70} & \underline{\textbf{84.24}} & \textbf{24.60} & \textbf{21.38} & \underline{\textbf{72.16}} & \textbf{33.11} & \textbf{29.20} \\
QwenVL2.5-7B & 96.20 & 94.47 & 92.40 & 91.50 & \textbf{90.82} & \textbf{84.69} & \textbf{91.32} & 50.48 & 49.52 & 77.42 & 56.03 & 50.42 \\
\rowcolor{gray!10}\ \ + \textsc{FOCUS} & 96.20 & \textbf{95.34} & \textbf{93.09} & 91.50 & 89.80 & 84.35 & 91.00 & \textbf{58.36} & \textbf{56.75} & \textbf{77.93} & \textbf{64.52} & \textbf{57.22} \\
LLaVA-OneVision-0.5B & 29.02 & 10.36 & 2.76 & 12.93 & 8.16 & 0.34 & 0.00 & \textbf{21.22} & 0.00 & 3.40 & 19.02 & 1.02 \\
\rowcolor{gray!10}\ \ + \textsc{FOCUS} & \underline{\textbf{29.36}} & \textbf{28.84} & \textbf{11.74} & \textbf{12.59} & \textbf{25.17} & \textbf{5.10} & \textbf{0.32} & 15.59 & \textbf{0.16} & \textbf{4.07} & \textbf{21.22} & \textbf{1.02} \\
LLaVA-OneVision-7B & 96.03 & 63.04 & 61.83 & 90.82 & \textbf{62.93} & \textbf{59.52} & 86.17 & 35.53 & 32.15 & 75.04 & 53.31 & 44.65 \\
\rowcolor{gray!10}\ \ + \textsc{FOCUS} & 96.03 & \textbf{69.43} & \textbf{67.88} & 90.82 & 58.84 & 56.12 & 86.17 & \underline{\textbf{45.82}} & \underline{\textbf{42.93}} & \textbf{74.70} & \textbf{69.95} & \textbf{55.86} \\
\bottomrule
\end{tabularx}
\caption{Performance across categories (Object, Attribute, Relation, Counting) with and without the proposed \textbf{\textsc{FOCUS}} decoding method. T: Text Score, I: Image Score, G: Group Score. Bold indicates improvement over the baseline. Underline highlights the largest gain within each model group for a given category.}
\label{tab:vismin_full}
\end{table*}

VisMin\cite{awal2024vismin} contains four semantic categories: Object, Attribute, Relation, and Counting. Image–text pairs differ only minimally within each category, which makes the benchmark highly sensitive to multi-image reasoning.
In Table~\ref{tab:vismin_full}, we present results with and without FOCUS.

\noindent\textbf{Object.} FOCUS shows clear gains in both the Image Score and the Group Score. These metrics are essential for evaluating cross-image understanding. Improvements appear in almost every model group except InternVL3-8B.

\noindent\textbf{Attribute.} In InternVL3-2B, FOCUS achieves nearly three times higher Group Score compared to the baseline. This result highlights its strengthened capability in distinguishing fine-grained visual properties across images.

\noindent\textbf{Relation.}
Llava-OV-7B baseline obtains an Image Score of 35.53, whereas FOCUS reaches 45.82. The Group Score also increases from 32.15 to 42.93. These improvements demonstrate significantly enhanced relational reasoning.

\noindent\textbf{Counting.}
FOCUS shows consistent improvements across all model groups in Counting. The InternVL3-2B model exhibits the largest performance increase.

\noindent\textbf{Summary.}
Overall, FOCUS consistently and robustly outperforms the baselines across all VisMin semantic categories. The method shows strong generalization in understanding each image while effectively mitigating cross-image information leakage.

\begin{figure*}[t]
  \centering
  \includegraphics[width=\textwidth]{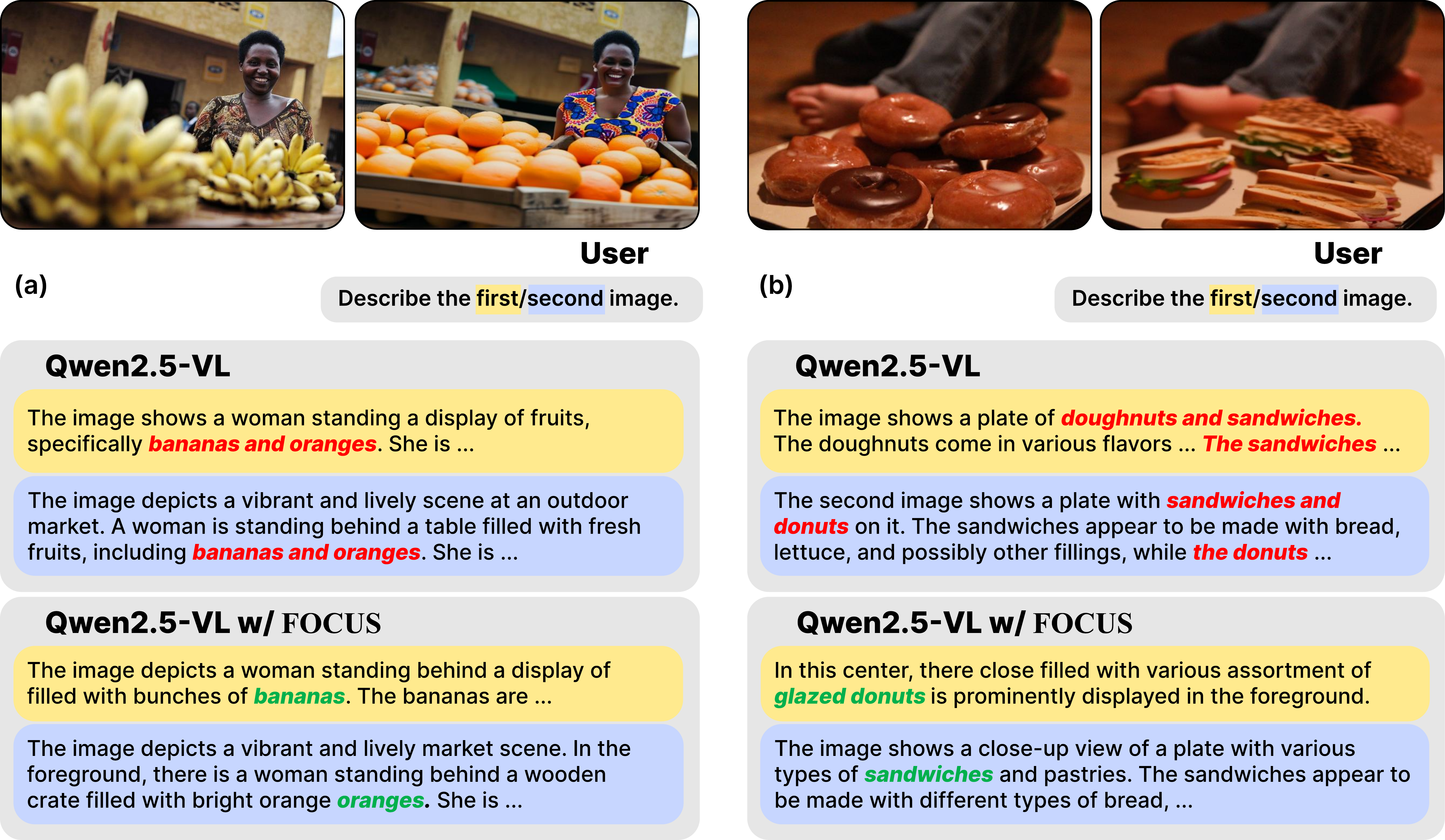}
  \caption{Qualitative samples using the Qwen2.5-VL 3B model. In both multi-image settings, the baseline decoding strategy often produces a mixed information of the other image not indicated by the question. On the other hand, FOCUS disentangles cross-image information effectively, resulting in better multi-image understanding. These examples illustrate that FOCUS helps generate image-specific responses while suprressing cross-image information leakage.}
  \label{fig:A_qualitative}
\end{figure*}
\begin{figure*}[t]
  \centering
  \includegraphics[width=\textwidth]{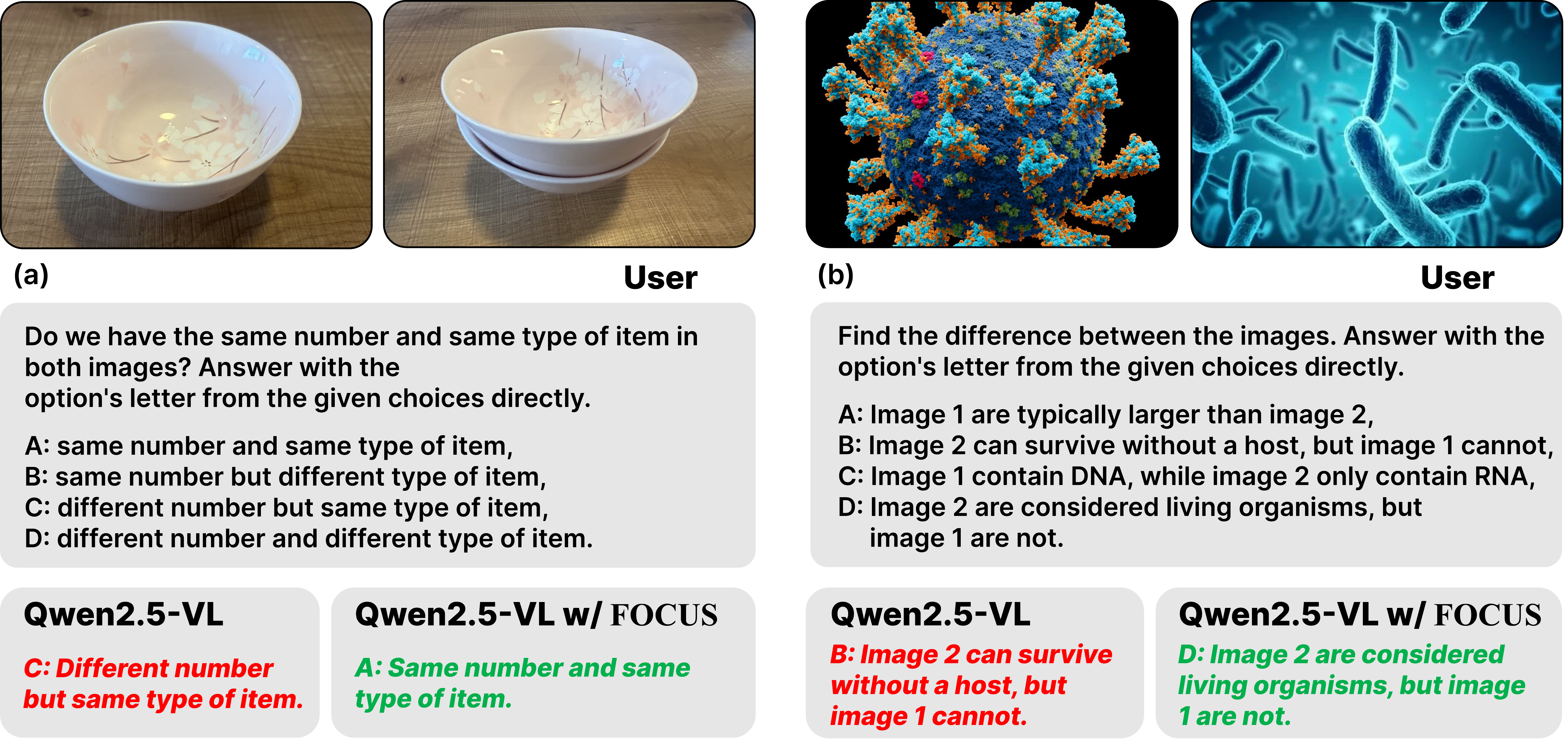}
  \caption{More qualitative samples using the Qwen2.5-VL 3B model. Examples are from the Mantis benchmark. Figure shows that baseline decoding confuses cross-image differences, while FOCUS correctly identifies image-specific information.}
  \label{fig:A_qualitative_more2}
\end{figure*}

\section{Additional Qualitative Results}
\label{app:additional_qual}

Figure~\ref{fig:A_qualitative} compares baseline decoding with FOCUS using examples from the VisMin benchmark. In (a), the baseline incorrectly mentions both bananas and oranges regardless of which image is queried. In contrast, FOCUS describes only bananas for the first image and only oranges for the second, showing a clear separation of visual content. In (b), the baseline mixes elements from both images, referring to doughnuts and sandwiches even when asked about one. FOCUS avoids this confusion and generates precise, image-specific descriptions. These examples show that baseline decoding fails to disentangle visual signals across multiple images, whereas FOCUS effectively suppresses cross-image information leakage.

Figure~\ref{fig:A_qualitative_more2} presents additional examples from the Mantis benchmark. In (a), the baseline misunderstands that the only difference between the two images is the number of items. FOCUS correctly identifies this distinction and provides the right answer. In (b), the two images depict bacteria and viruses. The model should respond accurately about the queried image, yet the baseline fails, while FOCUS answers correctly. These cases further highlight FOCUS's ability to maintain image-specific reasoning.

\section{Use of AI Assistants} We used AI assistants for polishing the manuscript and partially for code debugging. AI-generated images were used as input images in qualitative experiments to illustrate cross-image information leakage.

\end{document}